\documentclass[twoside,journal]{IEEEtran}
\usepackage[linesnumbered,ruled]{algorithm2e}

\usepackage{graphicx}

\usepackage{amsmath,amsfonts}
\usepackage{subcaption}
\usepackage{flushend}
\usepackage{makecell}
\usepackage{cite}

\begin{document}
\title{\LARGE Learning to Optimise Climate Sensor Placement using a Transformer
}

\author{Chen~Wang,~\IEEEmembership{Member,~IEEE,} 
        Victoria~Huang,
        Gang~Chen,~\IEEEmembership{Senior Member,~IEEE,} 
        Hui~Ma,~\IEEEmembership{Senior Member,~IEEE,} 
        Bryce~Chen, and Jochen Schmidt

\thanks{This work was supported by Agility Fund under Grant CDFP2319, administrated by the the National Institute of Water and Atmospheric Research, New Zealand (Corresponding author: Chen Wang.)}
        
\thanks{Chen~Wang, Victoria~Huang, Bryce Chen, and Jochen Schmidt are with the National Institute of Water and Atmospheric Research, Wellington 6021 New Zealand (e-mail: chen.wang@niwa.co.nz; victoria.huang@niwa.co.nz; bryce.chen@niwa.co.nz; jochen.schmidt@niwa.co.nz).}

\thanks{Hui~Ma, and~Gang~Chen are with the School of Engineering and Computer Science,  Victoria University of Wellington, Wellington 6041 New Zealand (e-mail: hui.ma@ecs.vuw.ac.nz; aaron.chen@ecs.vuw.ac.nz). }}




\maketitle

\begin{abstract} Optimal placement of climate sensors for environmental monitoring and disaster management presents a significant challenge due to its NP-hard complexity. Traditional sensor placement strategies have relied on exact, approximation, or heuristic methods, with heuristics being the most common due to their practicality. However, heuristics methods often depend heavily on expert knowledge, limiting their adaptability. Recent advances in deep learning offer a new avenue for enhancing heuristic algorithms either by generating them automatically or by guiding their search processes. In this paper, we propose a novel approach to sensor placement that leverages a Transformer-based network, trained through reinforcement learning (RL), to refine the search strategy of heuristic algorithms. By comparing our method against various heuristic-based strategies, we demonstrate its superior ability to generate high-quality solutions for the optimal sensor placement problem.
\end{abstract}

\section{Introduction} \label{sect:introduction}

Strategically placing climate sensors is crucial in climate research and forecasting for environmental monitoring and disaster management~\cite{andersson2022active}. By deploying sensors strategically, we can gather comprehensive and accurate information about climate factors such as precipitation and temperature. This data is essential for accurately predicting conditions in areas lacking sensor coverage, often achieved through spatial interpolation techniques \cite{andersson2022active,nguyen2020efficient}. Moreover, observations collected from climate sensors play a vital role in calibrating downscaled meteorology's climate models, such as ACCESS-S2 \cite{wedd2022access}. These models are utilized for weekly to seasonal forecasts, where adjustments are made to correct biases between observations and predictions. Consequently, the placement of climate sensors holds significant importance in the field of climate research and forecasting.

Determining optimal sensor locations is a class of optimization problems (henceforth referred to as \emph{Sensor Placement Problem}, shortly named as SPP), which poses a significant challenge to solve due to their NP-hard nature \cite{akbarzadeh2014}. Traditional methods for addressing sensor placement problems can be classified into exact methods~\cite{zhao2016new,xu2022multi,hart2008teva}, approximation methods~\cite{cheng2008relay,ma2015connectivity,le2019node}, and heuristics~\cite{das2020design, zaineldin2020improved,andersson2022active}. While exact methods provide optimal solutions, they struggle to scale to large environmental datasets due to high computational cost. Approximation methods can produce sub-optimal solutions, but these solutions may be far from the optimal \cite{MAZYAVKINA2021105400}. Heuristics, which are the most widely used approach, often provide satisfactory solutions within reasonable computational time frames. However, their development heavily relies on expert intuition and experience \cite{MAZYAVKINA2021105400}.

To improve heuristics based methods, deep learning (DL) has emerged as a promising approach to automatically generate heuristics or guide the search behaviours of heuristics ~\cite{wu2021learning,bengio2021machine}, leading to the creation of superior heuristics compared to those designed by humans \cite{cappart2020}. This successes is mainly due to two reasons: 1) the uniform data structure presents in a class of problem instances, with variations in data following a specific distribution, and 2) the capacity of DL models to detect patterns within a problem class using supervised learning/reinforcement learning (RL). For example, sequence-to-sequence models, such as transformers, has been incorporated into heuristic improvement for decision-making \cite{xin2021neurolkh,ma2021learning,zhang2020learning,wu2021learning}. The attention mechanism in transformers acts like a 'spotlight,' enabling the transformers to concentrate on the most relevant features and relationships of a class of problem instances, and is trained to optimise the performance of heuristic using supervised learning/RL.

Despite some recent successes in DL-based methods on improving heuristics for routing problems, such as the Traveling Salesman Problem (TSP)~\cite{kool2019attention,wu2021learning} and the Vehicle Routing Problem (VRP)~\cite{kool2019attention,nazari2018reinforcement,wu2021learning}. The problems investigated in the past are simplified benchmark problems in computer science domain, failing to address the complexity of real-world challenges, such as formulating a real-world optimisation problem (e.g., complex climate sensor placement problem) into a heuristic improvement problem, an absence of datasets or appropriate problem environments (e.g., simulators) for training DL models, and defining suitable objective functions (also named as reward functions in RL) for heuristic improvement. In other words, the application of using DL techniques to our problem remains unexplored.

The overall goal of this paper is to develop a DL-based sensor placement approach that optimise the locations of climate sensors via learning and improving heuristics on sensor placement. In contrast to traditional heuristics that rely on manually-designed search policies, our approach leverages RL to effectively guide the heuristic behaviour without extensive domain knowledge. We accomplish three primary contributions in this work:

\begin{enumerate}
\item We present an RL formulation on our optimal climate sensor placement problem as a heuristics improvement problem, where a RL policy is responsible for directing moving sensor to desired candidate sensor locations, enabling more effective search strategies. Moreover, we implement an environmental simulator designed specifically for the climate sensor placement problem. This simulator is capable of generating random problem instances represented by the state of the environment, performing actions to move sensors from one location to another, return rewards for every action and total reward over time.
\item We develop a DL-based sensor placement approach based on a transformer, which represent the RL policy and it is trained via an actor-critic algorithm. This algorithm enables this transformer-based policy to continuously learn and adapt based on the current state of the environment.
\item We conduct extensive experimental comparison of our method with other heuristic based approaches, ultimately demonstrating the effectiveness and superiority of our proposed approach in producing high-quality solutions.
\end{enumerate}

The remainder of this paper is structured as follows: Sect.~\ref{sect:literature_review} delves into related work concerning methods for solving optimization problems and provides essential preliminaries. Sect.~\ref{sect:problem formulation} details the formulation of our sensor placement problem. Sect.~\ref{sect: method} offers an overview of the method we propose. Sect.~V evaluates our method's effectiveness by contrasting it with recent algorithms. Sect.~\ref{sect:model_exploration} elucidates the underlying insights of the model. Finally, Sect.~\ref{sect:conclusion} concludes the paper and discusses potential avenues for future research.

\section{Literature Review} \label{sect:literature_review}

The challenge of optimal sensor placement for environmental monitoring and disaster management has been a subject of extensive research~\cite{andersson2022active}. A wide variety of methods have been proposed, each with its own advantages and shortcomings. These methods can be broadly grouped into three categories: exact methods, approximation methods, and heuristic methods. Recently, there has been a growing interest in learning improvement heuristics that leverage deep reinforcement learning (RL) to automatically discover effective improvement policies \cite{MAZYAVKINA2021105400}.

\subsection{Exact Methods}

Exact methods are designed to find the optimal solution to a problem, ensuring the best possible outcome. While there are various frameworks available, the branch and bound technique is one that is commonly employed in the design of these methods ~\cite{zhao2016new,xu2022multi,hart2008teva}. For example, the sensor placement problem in~\cite{zhao2016new} was formulated as a mixed integer convex programming in water sensor networks. Through convex relaxation, a branch and bound algorithm was proposed to find the global optimum. Similarly, a toolkit was developed in~\cite{hart2008teva} to combine general purpose heuristics with bounding algorithms and integer programming.

However, due to their high computational complexity, exact methods are usually limited to small problem instances. Despite their limitations, exact methods are a valuable benchmark for assessing the performance of other, more computationally feasible methods.

\subsection{Approximation Methods}

Approximation methods provide a balance between computational feasibility and solution quality \cite{carr2006robust,ma2015connectivity,le2019node,cheng2008relay}. These methods, including linear programming relaxations and local search algorithms \cite{carr2006robust,ma2015connectivity}, do not guarantee an optimal solution but offer solutions within a known range from the optimum. This makes them a more practical choice for large problem instances. For example, an approximation algorithm with a constant approximation ratio based on a divide and conquer technique named partitioning and shifting was proposed in~\cite{le2019node} with the goal of maximizing the sensor coverage. Similarly, approximation algorithms have also been designed in~\cite{cheng2008relay}. In~\cite{ma2015connectivity}, a local search approximation algorithm was proposed where sensors were allocated into groups and local search was applied within each group to find the sensor locations. 

However, one of the main challenges with approximation methods is that they can yield solutions that deviate significantly from the optimal solution, especially in cases where local optima are far from the global optimum \cite{addis2005local}.

\subsection{Heuristic Methods}

Heuristic methods use rules of thumb or educated guesses to find satisfactory solutions within a reasonable time frame \cite{andersson2022active,das2020design,zaineldin2020improved, harizan2019coverage,wang2018pso,yarinezhad2020sensor,zhang2021novel}. They are particularly useful when dealing with complex problems where exact or approximation methods are not practical.

A greedy algorithm was proposed in~\cite{andersson2022active} to iteratively select sensor locations with the highest ranking score calculated by different measurements, such as the network coverage and the mean absolute error \cite{bromwich2004strong}. Another popular heuristic method is Genetic Algorithms (GA), which has been used to deploy a minimum number of sensor nodes, while maximizing the coverage~\cite{das2020design, zaineldin2020improved, harizan2019coverage}. Other heuristics such as Particle Swarm Optimization (PSO)~\cite{wang2018pso,yarinezhad2020sensor}, and Simulated Annealing (SA)~\cite{zhang2021novel}, have also been widely used in sensor placement.

The above heuristic methods rely heavily on expert intuition and experience for designing the heuristic rules, which makes the development of heuristics a complex and time-consuming process. While heuristics typically function without assured optimal results, in certain circumstances, it is possible to establish a worst-case performance bound. This boundary delineates the maximum potential deviation of the solution from the optimal result.

\subsection{Learning Improvement Heuristics}

Recognizing the limitations of traditional heuristic methods, researchers have started to explore the idea of learning based approaches to address TSP~\cite{khalil2017learning, vinyals2015pointer, deudon2018learning} and VRP~\cite{kool2019attention, nazari2018reinforcement}. All these works learned heuristics to construct a complete solution directly from scratch. However, when the number of decision variables increases, it can be challenging to construct solutions directly~\cite{chen2019learning}.

Meanwhile, learning improvement heuristics has been proposed \cite{chen2019learning,lu2020learning,wu2021learning}. Rather than constructing solutions from scratch, improvement heuristics are learned to iteratively improve a given solution. For example, NeuReWriter~\cite{chen2019learning} formulated the VRP as a rewriting problem and learned two policies: a region-picking policy and a rule-picking policy. Given a state, the region-picking policy first picks a region (i.e., the partial solution) and the rule-picking policy picks a rewriting rule to apply to the selected region to generate an improved solution. Similarly, a learning improvement heuristic for capacitated VRP was proposed in~\cite{lu2020learning}. Started with a feasible solution, the algorithm iteratively updates the current solution with an improvement operation selected by RL. Meanwhile, the algorithm also perturbs the current solution with a rule-based operation to partially or completely destroy and reconstruct a solution whenever a local minimum is reached. 

Nevertheless, existing learning based methods predominantly focus on routing problems~\cite{kool2019attention,wu2021learning,nazari2018reinforcement,chen2019learning,lu2020learning} while the potential for applying learning based techniques to other challenging problems (e.g., sensor placement) remains largely unexplored. 


\subsection{Preliminaries}\label{sect:premilinary}
\subsubsection{Attention models}

Attention-based models have emerged as an influential component in sequence modeling tasks, especially,  nature language processing problems. Bahdanau and his team introduced an intuitive but potent form of attention, named Additive-Attention \cite{bahdanau2014neural}, which highlights the significance of certain words in reference to an external query. This principle was later expanded by Vaswani et al. \cite{vaswani2017attention}, who proposed the concept of Self-Attention or Multiplicative-Attention. Unlike basic attention, self-attention takes into account the interaction between words within the same sentence.

Eq.~\ref{eq:attetion_general} mathematically illustrates self-attention mechanism. The mechanism begins by creating three distinct vectors from the input embeddings, named $Query (Q)$, $Key (K)$, and $Value (V)$. The initial stage comprises the calculation of the dot product between the Q and K vectors, which results in an attention map. In this map, related entries score high, while unrelated ones receive lower scores. This map is then scaled by the square root of the embedding dimension ($d_h$), followed by processing through a softmax function to form a probability matrix. The multiplication of this probability matrix with the V vector generates the final output, emphasizing the elements of focus.

\begin{equation}
\label{eq:attetion_general}
\mathbf{A T T}\left(Q, K, V \right)=V \cdot \operatorname{softmax}\left(\frac{K^T Q_{}}{\sqrt{d_h}}\right)
\end{equation}


\section{Problem Formulation}\label{sect:problem formulation}

\subsection{SPP formulation}
In our SPP, we consider the possible placements of a set of \textit{sensors} for climate station network. Formally, a \textit{sensor} is considered as a tuple $s_i = (p_i, z_i)$ where $p_i$ is the location of the $i$th sensor, and $z_i$ is an observed value of a target environmental variable at a site location $p_i$. Location $p_i$ is also considered as a tuple $p_i = (x_i, y_i)$, and $x_i$ and $y_i$ are latitude and longitude of the location $p_i$. 

The set of all sensors in the climate station network joint form a \textit{sensor network}. Formally, a \textit{sensor network} is considered as a set of tuple $\mathcal{S} = \{s_1, s_2, \dots, s_i, \dots, s_n\}$, made of a group of interconnected sensors that monitor environmental variables, with $n$ being the total number of sensors to be installed. For any arbitrary sensor network $\mathcal{S}$, the set of site locations of a sensor network $\mathcal{S}$ is denoted as $\mathcal{P} = \{p_1, p_2, \dots, p_i, \dots, p_n\}$.

\begin{figure}[h]
    \centering
    \scalebox{0.55}{\includegraphics{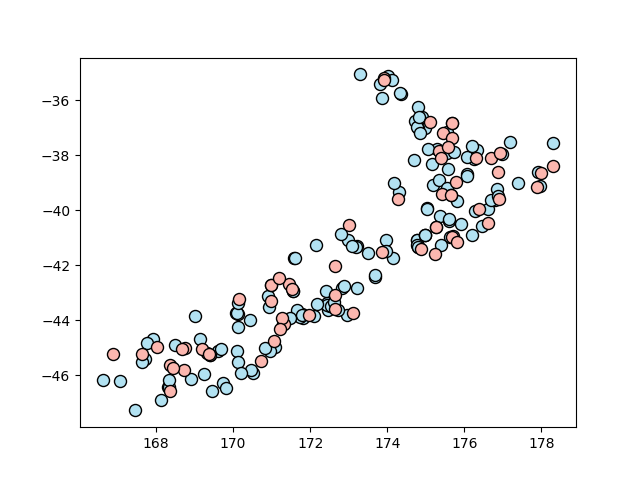}}
    \caption{An example of a full sensor site location set $\mathbb{P}$  }
    \label{Fig:NZsensors}
\end{figure}

We consider $m$ candidate climate site locations, defined as $\mathcal{Q} = \{p_{n+1}, \dots, p_k, \dots, p_{n+m}\}$ with $\mathcal{Q} \cap \mathcal{P} = \emptyset $, that can be used to relocate sensors from $S$. The full sensor site location set considered in SPP is denoted as $\mathbb{P}  = \mathcal{P} \cup \mathcal{Q} = \{p_1, p_2, \dots, p_n, p_{n+1}, \dots, p_{n+m}\}$. Fig.~\ref{Fig:NZsensors} shows an example of a full climate sensor site location set  $\mathbb{P}$  on a regular 5km grid covering New Zealand, with an sensor site $p_{i} \in \mathcal{P}$ in red, and candidate climate sites $p_{k} \in \mathcal{Q}$ in blue.  

In practice, we often rely on a spatial interpolation technique to estimate the target environmental variable, $\hat{z}_j$, at any geographic location $p^{\prime}_j$ on the map \cite{andersson2022active}. Often, we are interested in estimating the target variable at a set of geographic locations $\mathcal{P}^{\prime} = \{p_{1}^{\prime}, p_{2}^{\prime}, \dots, p_{j}^{\prime}, \dots, p_{q}^{\prime}\}$ with $\mathcal{P^{\prime}} \cap \mathcal{P} = \emptyset $. Herein we consider a simple but effective technique, i.e., Inverse Distance Weighting (IDW) technique \cite{lu2008adaptive}, to estimate $\hat{z}_j$ using $n$ observations from a sensor network $\mathcal{S}$.


\begin{equation}
\label{eq:interpolation}
\hat{z}_j = \frac{\sum\limits_{i=1}^n w_i z_i}{\sum\limits_{i=1}^n w_i}
\end{equation}

\noindent where $w_i$ is a weight assigned to the $i^{th}$ sensor $s_i$ in a sensor network $\mathcal{S}$, given by $w_i = \frac{1}{d_i}$, and $d_i = || p_i - p^{\prime}_j||_2$ is an Euclidean distance between the $i$th sensor and the location $p_j$, determining the degree of influence of the distance on the weight $w_i$.

The problem investigated in this paper aims to find an optimized sensor network $\mathcal{S}^{\star}$, where some sensors in  sensor network $\mathcal{S}$ are strategically reallocated in different but available places in $\mathbb{P}$. Specifically, we aims to minimize the Mean Absolute Error (MAE) over all estimated values for $q$ geographic positions $P^{\prime}$ as follows:

\begin{flalign} \label{eqn:mae}
    MAE(\mathcal{S}) 
    &= MAE(\mathcal{S} ~|~ \mathcal{P}^{\prime}) \nonumber \\ 
    &= \frac{1}{q}\sum_{j=1}^{q} |z_j-\hat{z}_j|,\nonumber \\
    &\qquad \forall~ z_{j} \in \mathcal{S}
\end{flalign}

\noindent where $z_j$ is the ground truth, i.e., actual value of the $j^{th}$ estimation, and $\hat{z}_j$ is the predicted value of the $j^{th}$ observation. A lower $MAE$ indicates a better prediction and a better sensor network.

\subsection{MDP formulation of the SPP}
We formulate the Markov Decision Process (MDP) as follows:
\begin{enumerate}
\item State $\mathcal{ST}^t$: the state $\mathcal{ST}^t$ represents a problem instance at time step $t$, i.e., a sequence of selected sensor locations and a sequence of candidate locations where sensors can move to. The initial state $\mathcal{ST}^0$ is an initial solution that we aim to improve using a heuristic $\mathcal{H}$. For example, $\mathcal{ST}^t = [p_1, p_2, \dots, p_n, | p_{n+1}, \dots, p_{n+m} ]$. Each position on the left side of $|$ corresponds to a sensor location in $\mathcal{P}$, while the right side corresponds to the remaining available candidate locations in $\mathcal{Q}$. Note that $|$ is just displayed for the courtesy of the reader, not part of the state. Moreover, a sensor network $\mathcal{S}^t$ can be decoded from the state of the sensor network $\mathcal{ST}^t$ at that specific time step $t$. We can noted it as $\mathcal{S}^t \Longleftarrow \mathcal{ST}^t$.

\item Action $\mathcal{A}^t$: any action $\mathcal{A}^t$ corresponds to selecting a pair of locations $(p_a, p_b)$ and moving the sensor from position $p_a \in \mathcal{P}$ to position $ p_b\in \mathcal{Q}$.

\begin{figure}[!hbt]
	\centering
	\includegraphics[width=.38\textwidth]{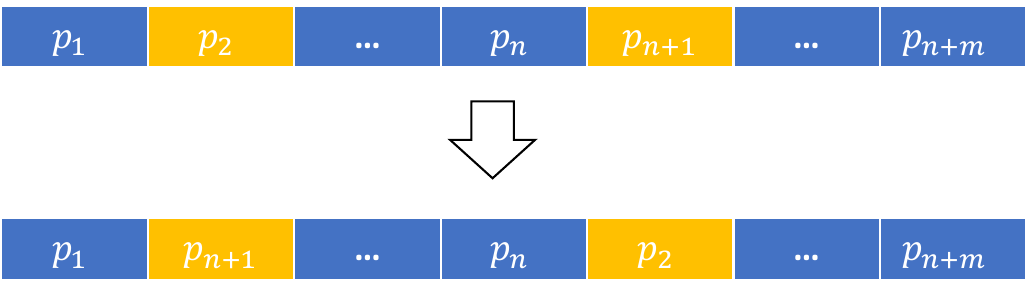}
	\caption{Moving a sensor from position $p_2$ to position $ p_{n+1}$.} 
\label{fig:swap}
\end{figure}

\item Transition $\mathcal{T}$: the next state $\mathcal{ST}^{t+1}$ is obtained in a deterministic manner from $\mathcal{ST}^t$ by performing an action $\mathcal{A}^t$, i.e., $\mathcal{ST}_{t+1} = \mathcal{T}(\mathcal{ST}^t, \mathcal{A}^t)$.

\item Reward $\mathcal{R}^t$: the reward function $\mathcal{R}^t$ is designed to best improve the initial solution within $T$ steps as follows:

\begin{equation}
\resizebox{0.9\hsize}{!}{$
\begin{aligned}
\label{eq:reward}
\mathcal{R}^t &= \mathcal{R}^t(\mathcal{ST}^t, \mathcal{A}^t, \mathcal{ST}^{t+1}) \\
              &= MAE_{best}(\mathcal{ST}^{t}) - \min\{MAE_{best}(\mathcal{ST}^t),
MAE(\mathcal{ST}^{t+1})\}
\end{aligned}
$}
\end{equation}

Where $\mathcal{ST}^{t}$ is the best solution found up to step $t$ and is updated when $\mathcal{ST}^{t+1}$ is a better solution. The reward $\mathcal{R}^t$ is positive only when a better solution is found; otherwise, it equals 0. The objective is to maximize the cumulative reward $G_T = \sum_{t=0}^{T-1} \gamma^t \mathcal{R}^t$, where $\gamma$ is the discount factor. This discount factor, $\gamma$, typically lies in the range of 0 to 1, which is used to balance the relative importance of immediate rewards versus future rewards. A value closer to 0 makes the model short-sighted, focusing on immediate rewards, while a value closer to 1 encourages the model to consider long-term rewards. 
\end{enumerate}

\subsection{Simulation of the SPP}\label{subs: simulation}

To address the lack of datasets or suitable environments (e.g., simulators) for training deep learning (DL) models, we generate synthetic problem instances of the SPP to effectively train our transformer-based policy. For each problem instance, every sensor location — latitude $x_i$ and longitude $y_i$—of $n+m$ sensors in $\mathcal{ST}^0$ are sampled randomly across a specified area of interest on the map. Following this, we employ an IDW model $\mathcal{D}$ to estimate the target observation value at each sampled location. The IDW model is constructed and trained using real-world sensor observation data. This model is chosen for its efficiency in spatial interpolation and estimation. For more details, please refer to simulation settings in Sect.~\ref{sect: experiment}

\section{A Novel SPP Approach Based on A Transformer}\label{sect: method}

\begin{figure*}[!hbt]
	\centering
	\includegraphics[width=.85\textwidth]{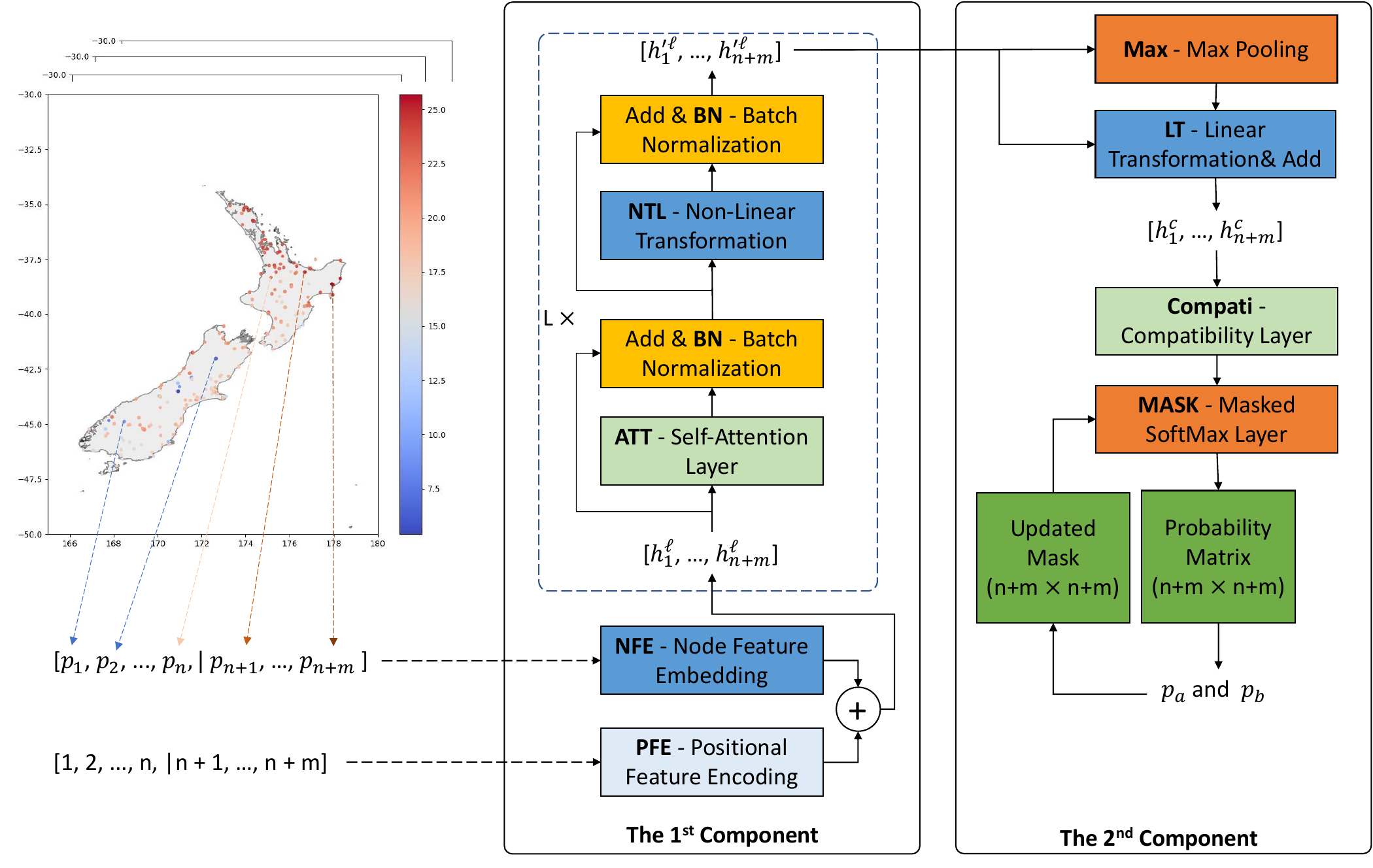}
	\caption{Our policy network architecture} 
\label{fig:policy}
\end{figure*}

Our policy network is composed of two main components, as illustrated in Figure~\ref{fig:policy}. The first component learns a sequence embedding for sensor locations (i.e., the state $\mathcal{ST}^t$) via a $L$ stacked encoder with self-attention. The second component focuses on computing the compatibility between sensor location pairs (i.e., the action $\mathcal{A}^t$). Using self-attention, it produces a probability matrix, where each element represents the likelihood of selecting the corresponding sensor location pair to guide an action. In this matrix, each row represents a specific sensor at its current location, while each column represents a potential destination location for that sensor. Therefore, the entry in the $a$-th row and $b$-th column corresponds to the probability of moving the sensor at location $p_a$ to location $p_b$. In essence, this matrix captures the network's strategy for reallocating sensors to improve the quality of the sensor network. The two components of the policy network are defined in the underlying Eq.~\ref{eq:transformer_part1} \footnote{The time step t is omitted here for better readability} and  Eq.~\ref{eq:transformer_part2}, respectively, and also explained below.

\begin{equation}
\begin{aligned}
\label{eq:transformer_part1}
[h_1^{0}, ..., h_{n+m}^{0}] & = \mathbf{NFE}(\mathcal{ST}) + \mathbf{PFE}(\mathcal{ST})\\
H^{0} & = [h_1^{0}, ..., h_{n+m}^{0}] \\
H^{\ell} & =\mathbf{B N}^{\ell}\left(H^{(\ell-1)}+\mathbf{A T T}^{\ell}\left(H^{\ell-1}\right)\right) \\
H^{\prime \ell} & =\mathbf{B N}^{\ell}\left(H^{\ell}+\mathbf{NLT}^{\ell}\left(H^{\ell}\right)\right), \ell=1, \ldots, L
\end{aligned}
\end{equation}

\begin{equation}
\begin{aligned}
\label{eq:transformer_part2}
H^{c} & = \mathbf{LT} \left(\mathbf{Max}(H^{\prime L}) \right) + \mathbf{LT} \left(H^{\prime L}\right)\\
M & = \mathbf{Compati}\left(H^{c}\right)\\
PR & = softmax \left(\mathbf{MASK}\left(M \right)\right)
\end{aligned}
\end{equation}

\textbf{NFE} - Node Feature Embedding: We use linear transformation to project every sensor location, i.e., $p_i = (x_i, y_i)$, from state $\mathcal{ST}$ into a $d_{h}$-dimension embedding using a linear transformation.

\textbf{PFE} - Position Feature Embedding: Let $i$ denote the sequence position of a sensor in $\mathcal{ST}$, and let $d=1,2,...,d_h$ denote the dimension index. The functions $\lfloor \cdot \rfloor$ and $\text{mod}$ represent the floor and modulo operations, respectively. Following the position feature embedding in ~\cite{vaswani2017attention}, Eq.~\ref{eq:position_embedding} employs sinusoidal positional encoding to represent the position of each sensor position in the sequence. Note that it is important to differentiate between the terms ``sensor location'' and ``sensor position'' in this context. ``Sensor location'' refers to the geographic coordinates (latitude and longitude) of the sensor, whereas ``sensor position'' refers to the index of the sensor in the sequence.

\begin{equation}
\label{eq:position_embedding}
\begin{aligned}
g(i, d)=\left\{\begin{array}{l}
\sin (i / 10000^{\frac{\lfloor d/2 \rfloor}{d_h}}), \text { if } d \text { is even } \\
\cos (i / 10000^{\frac{\lfloor d/2\rfloor  }{d_h}}),  \text { if } d \text { is odd }
\end{array}\right.
\end{aligned}
\end{equation}

\textbf{ATT} - Self-attention: In alignment with Eq.~\ref{eq:attetion_general}, the self-attention mechanism can be applied to an input matrix $H^{(l-1)} = [h_1^{(l-1)}, ..., h_{n+m}^{(l-1)}]$, which is produced by \textbf{NFE} and \textbf{PFE}. The self-attention can thus be expressed by the following equation:

\begin{equation}
\label{eq:attetion}
\mathbf{A T T}^{\ell}\left(H^{\ell-1}\right)=V^{\ell} \cdot \operatorname{softmax}\left(\frac{(K^{\ell})^T Q^{\ell}}{\sqrt{d_h}}\right)
\end{equation}

\noindent where the query, key, and value matrices of $H^{{\ell}-1}$ are given by $Q^{\ell}=W_q^{{\ell}}H^{{\ell}-1}$, $K^{\ell}=W_k^{{\ell}}H^{{\ell}-1}$, and $V^{\ell}=W_v^{{\ell}}H^{{\ell}-1}$, respectively. $W_q^{{\ell}}$, $W_k^{{\ell}}$, and $W_v^{{\ell}}$ are the weight matrices to be trained.

\textbf{NTL} and \textbf{BN} - Non-Linear Transformation and Batch Normalization: NTL perform a weighted sum of inputs, add a bias term, and apply an activation function to produce an output. NTL can be computationally expensive and prone to overfitting. In Eq.~\ref{eq:transformer_part1}, we further incorporate Batch normalization and Skip connection techniques, which help to stabilize the training process and mitigate overfitting.

\textbf{Max} and \textbf{LT} - Max-pooling and Linear Transformation: In Eq.~\ref{eq:transformer_part2}, we combine the embeddings using max-pooling and subsequently enhance the resulting embedding by transforming $H^L$ into $H^c$ via Linear Transformation. This design effectively integrates the global information of an instance into its corresponding embedding.

\textbf{Compati} - Compatibility Matrix:
Compatibility has proven to be effective in representing connections among words within sentences. Similarly, this concept is applied to predict sensor location pairs in a sensor network for a moving operator. Given the embeddings $H^c = [h_1^c, ..., h_{n+m}^c]$, we compute the dot product between the query matrix $Q_{c}$ and key matrix $K_{c}$, as seen in Eq~\ref{eq:compatibility}. Both $K_{c}$ and $Q_{c}$ are derived in a manner akin to $(K^{\ell})^T$ and $ Q^{\ell}$ in Eq.~\ref{eq:attetion}. Each element $M_{a,b}$ in the compatibility matrix $M$ signifies the score associated with selecting each sensor location pair $(p_a, p_b)$.

\begin{equation}
\label{eq:compatibility}
M = Compati(H^{c}) = K_{c}^T Q_{c}
\end{equation}

\textbf{MASK} - Mask matrix: We introduce a mask to the compatibility matrix, as demonstrated in Eq.~\ref{eq:mask}. The diagonal elements are masked as they hold no meaningful value for position pair selection, and a tanh function is employed to confine the compatibility matrix values within the range $[-C, C]$. Therefore, the entry $pr_{a, b}$ in $PR$ represents the probability of moving a sensor from $p_a$ to $p_b$.

\begin{equation}
\begin{aligned}
\label{eq:mask}
MASK(M) & = \begin{cases} C \cdot \tanh \left(M\right), & \text { if } a \neq b \\
-\infty, & \text { if } a=b\end{cases} \\
\end{aligned}
\end{equation}

\subsection{Training the Transformer Model}

The continuous n-step actor-ccritic algorithm is an advanced reinforcement learning (RL) method that combines the advantages of both actor-critic and n-step bootstrapping \cite{mnih2016asynchronous,wu2021learning}. This algorithm provides a flexible framework for optimizing policies and value functions in a RL setting, making it suitable for training a continuing (also called non-episodic) RL task. Our SPP is formulated as a continuing RL task, since the action performed on the environment is ongoing without a definite end, requiring continuously learning and optimizing.

As shown in \textbf{Algorithm~\ref{alg:n_step_actor_critic}}, the algorithm takes as input the number of timesteps per update ($T_n$) and the maximum episode length ($T$). It initializes the policy and value function parameters, iterating over several epochs. In each epoch, $\mathbf{M}$ problem instances are initiated. For each batch drawn from these instances, actions are selected based on the current state and policy. The algorithm observes the consequent state and reward. For every $T_n$ steps, the future rewards are predicted using the value function. The algorithm then goes back through each of the last $T_n$ timesteps, calculating a TD-error (Temporal-Difference error), which reflects the difference between the predicted and current value estimates. This error is then used to compute gradients for updating the policy and value function parameters. The average of these gradients across all instances in the batch and the last $T_n$ timesteps is used to adjust these parameters. After updating, the process repeats for the next batch. The loop continues until termination criteria are met, resulting in the final policy and value function parameters. This algorithm can effectively learns continuous control policies, effectively managing the exploration-exploitation trade-off.

\begin{algorithm}
\SetAlgoLined
\KwIn{Number of timesteps per update $T_{n}$, Max episode length $T$}
\KwOut{Updated policy $\pi_{\theta}$ (Actor) and value function $V_{\phi}$ (Critic) parameters $\theta$ and $\phi$}
Initialize policy $\pi_{\theta}$ (Actor) and value function $V_{\phi}$ (Critic) parameters $\theta$ and $\phi$\;
\For{each epoch}{
    Initialize $\mathbf{M}$ problem instances and $t = 0$\;
    \For{each batch $B$ sampled from $\mathbf{M}$}{
        \Repeat{$t < T$}{
            Select action $\mathcal{A}^t \sim \pi_{\theta}(\cdot|\mathcal{ST}^{t})$\;
            Observe next state $\mathcal{ST}^{t+1}$ and reward $\mathcal{R}^t$\;
            $t \gets t + 1$, $d\theta \leftarrow 0$, $d\phi \leftarrow 0$\;
            \If{$t \mod T_{n} = 0 $}{
                $\hat{R} \leftarrow V_{\phi} (\mathcal{ST}^{t})$\;
                \For{$i \in \{t-1, \dots, t-T_{n}\}$}{
                    $\hat{R} \leftarrow \mathcal{R}^i+\gamma \hat{R}$\;
                    $\delta \leftarrow \hat{R} - V_{\phi} (\mathcal{ST}^{i})$\;
                    
                    $d\theta \leftarrow d\theta + \sum_{|B|} \delta \nabla \log \pi_{\theta} (\mathcal{A}^t|\mathcal{ST}^t)$\;
                    
                    $d\phi \leftarrow d\phi + \sum_{|B|} \delta \nabla \log V_{\phi} (\mathcal{ST}^t)$\;
                }
                update $\theta$ by $\frac{d\theta}{|B|T_n}$ \;
                update $\phi$ by $\frac{d\phi}{|B|T_n}$\;
            }
        }
    }
}
\caption{Continuous n-Step Actor-Critic}
\label{alg:n_step_actor_critic}
\end{algorithm}

\section{Experiment}\label{sect: experiment}
To evaluate the performance of our proposed method, we conduct experimental evaluations using a simulator based on real-world data, comparing with serveral baseline methods.

\begin{figure}[!hbt]
\small
\centering
	\begin{subfigure}{0.65\columnwidth} 
		\includegraphics[width=\textwidth]{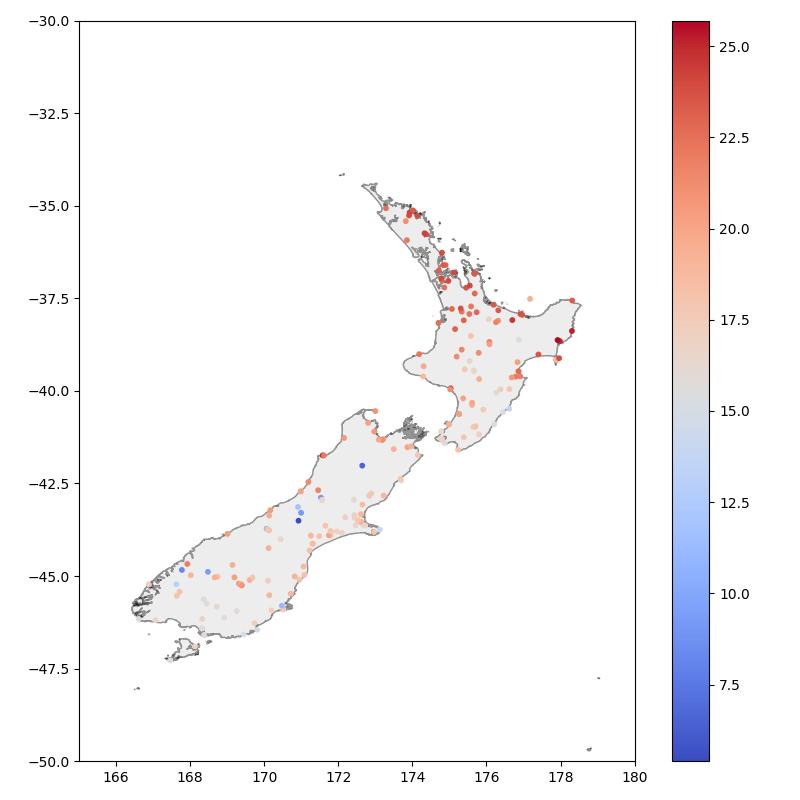}
		\caption{Training set (i.e., 206 sensor locations)} 
	\end{subfigure}
	\begin{subfigure}{0.65\columnwidth} 
		\includegraphics[width=\textwidth]{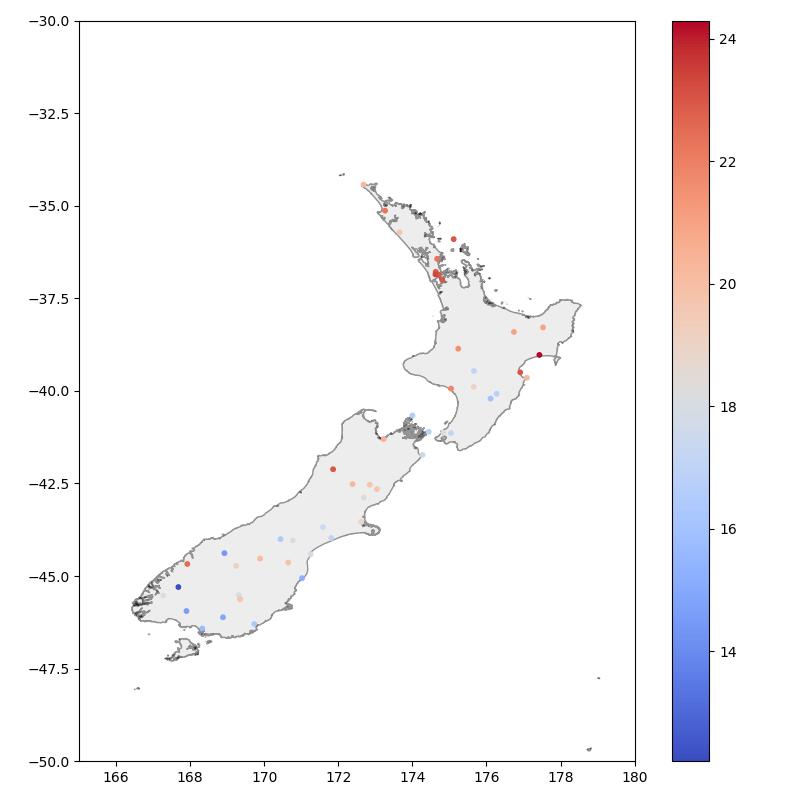}
		\caption{Testing set (i.e., 52 sensor locations)} 
	\end{subfigure}
\caption{Daily maximum temperature map of New Zealand}
\label{fig:dataset}
\end{figure}

\textbf{Simulation settings}. 
In the simulation, an IDW model is learnt from the observation data from the National Climate Database (CliDB)\footnote{CliDB is an online climate data platform provided by NIWA, New Zealand. For more information, visit https://cliflo.niwa.co.nz/.}, maintained by the National Institute of Water and Atmospheric Research (NIWA) in New Zealand. We use daily maximum temperature as an example of target climate varaible in this study. The daily maximum temperature was recorded from 258 NIWA temperature sensors (including CWS - NIWA Compact Weather Station and EWS - NIWA Electronic Weather Station). The 258 NIWA temperature sensors are partitioned into two distinct sets for training and testing purposes. Specifically, a random subset comprising 20\% of these observations, equating to 52 sensor locations, corresponds to the testing set (i.e., $\mathcal{P}^{\prime}$, as defined in Sect.~\ref{sect:problem formulation}). The remaining 80\%, consisting of 206 sensor locations, forms the training set. Fig.~\ref{fig:dataset}(a) and Fig.~\ref{fig:dataset}(b) provides a visual representation of the two sets of daily maximum temperatures.

For each training epoch, we generate 5120 problem instances, each comprising a sensor network with 206 random sensor locations within New Zealand's boundaries, using EfrainMaps' Shapefiles \footnote{EfrainMaps supplies ESRI format shapefiles (*.shp) for various countries and worldwide. For more information, visit https://www.efrainmaps.es/}. These locations are assigned ground truth values for maximum temperature, generated via an Inverse Distance Weighting (IDW) model trained on our initial dataset, as shown in Fig.~\ref{fig:dataset}(a). Through this approach, we create a large, diverse array of problem instances, improving the robustness and generalizability of our policy.


\textbf{Parameter Settings}.
In each training epoch, all the pre-generated 5120 problem instances are divided into ten distinct batches for training our transformer model. To reduce the operation cost, 60 sensors are selected from 205 sensor locations. As previously mentioned, improvement heuristics are modeled as a continuing RL task. Nevertheless, the agent is trained for a modest step limit ($T$ = 200). In our following, we will demonstrate that the trained policies exhibit strong generalization capabilities in unseen initial solutions and with significantly larger step limits (i.e., 1000 steps) in the testing phase. The discount factor, $\gamma$, is assigned a value of 0.99, while the n-step return parameter, n, is set to 4.

Training is performed over 200 epochs with an initial learning rate of $10^{-4}$ and a decay rate of $0.99$ applied to both the actor and critic networks, following reported values in \cite{wu2021learning}. Leveraging the computational power of 4 Tesla A100 GPUs, the average training time for each epoch is around 30 minutes. The Pytorch-based source code and pre-trained models related to this study can be accessed on Gitlab \footnote{The source code and instructions can be obtained from https://git.niwa.co.nz/rl-group/spp-transformer-ac}.

\textbf{Competing algorithms}.
The DL-based methods we propose and evaluate in this study include the \textit{Tran-mask swap} and \textit{Tran-swap}. The former, `Tran-mask swap', include the mask introduced in Eq.~\ref{eq:mask}. The latter, only permits swap actions performed over two distinct locations from $\mathcal{Q}$ and $\mathcal{P}$, respectively. 

In our experiment, we compare our proposed method with two baseline algorithms, Stochastic Search and Context Distance Search \cite{andersson2022active}. Stochastic Search randomly selects and moves a pair of sensor locations in each of its 1,000 iterations, ultimately returning the best solution with the lowest Mean Absolute Error (MAE). Conversely, Context Distance Search employs a heuristic strategy aiming to maximize the collective distances amongst sensors. This is crucial as it broadens coverage and minimizes redundancy, potentially leading to more efficient data collection and better environmental monitoring. This algorithm exhaustively traverses all possible configurations to find the solution with the maximum sensor separation. Comparing these baseline algorithms with our method provides valuable insights into each approach's relative strengths and weaknesses in tackling SPP.

All methods, including two baselines, are tested over 1000 randomly generated problem instances. These instances represent different scenarios of the sensor placement solutions, allowing us to assess the performance of the methods in a variety of situations to understand their general applicability.

\subsection{Performance Comparison}

\begin{figure*}[!hbt]
\small
\centering
	\begin{subfigure}{0.95\columnwidth} 
		\includegraphics[width=\textwidth]{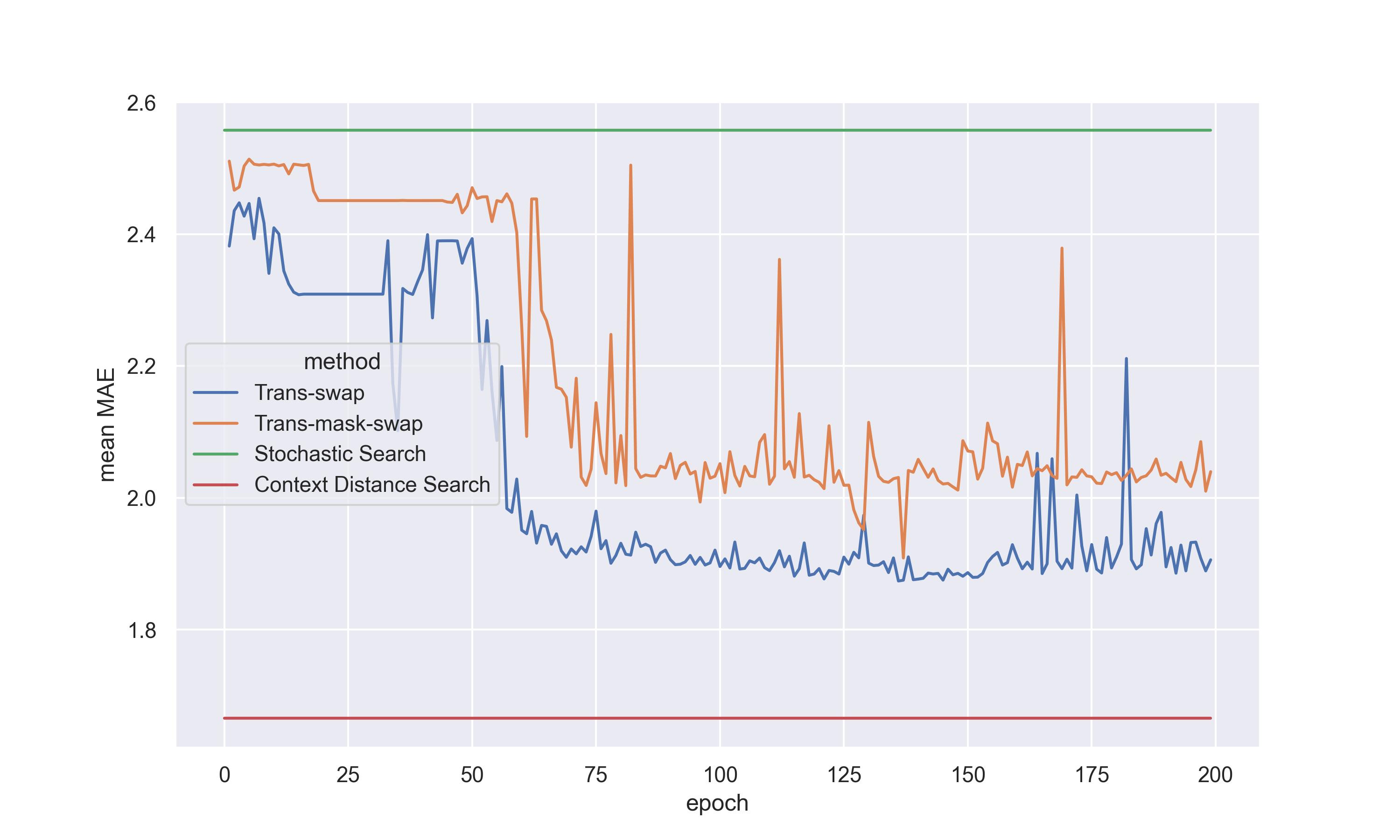}
		\caption{Mean of the mean MAE values on 20 \% testing instances} 
	\end{subfigure}
	\begin{subfigure}{0.95\columnwidth} 
		\includegraphics[width=\textwidth]{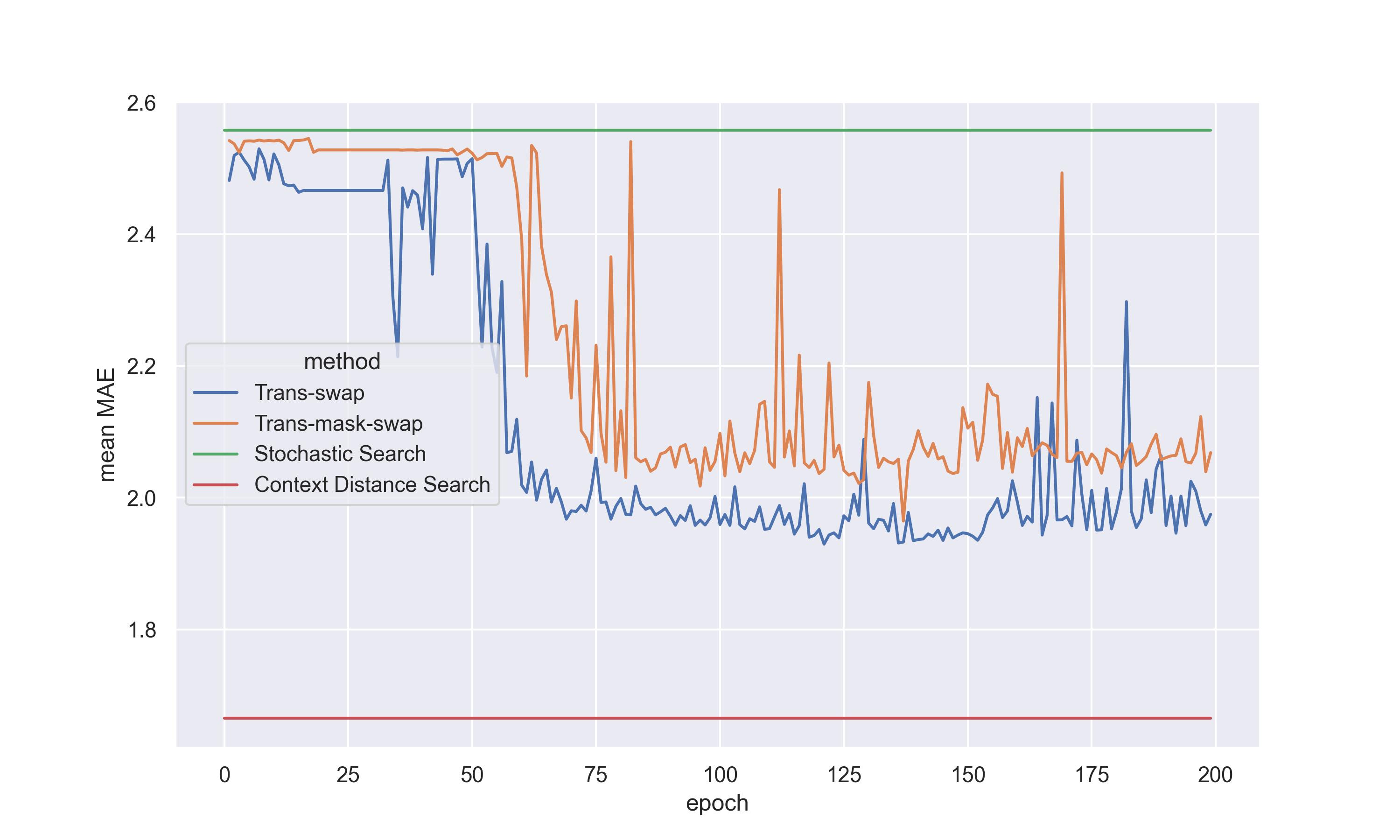}
		\caption{Mean of the meaan MAE values on 40 \% testing instances} 
	\end{subfigure}
 	\begin{subfigure}{0.95\columnwidth} 
		\includegraphics[width=\textwidth]{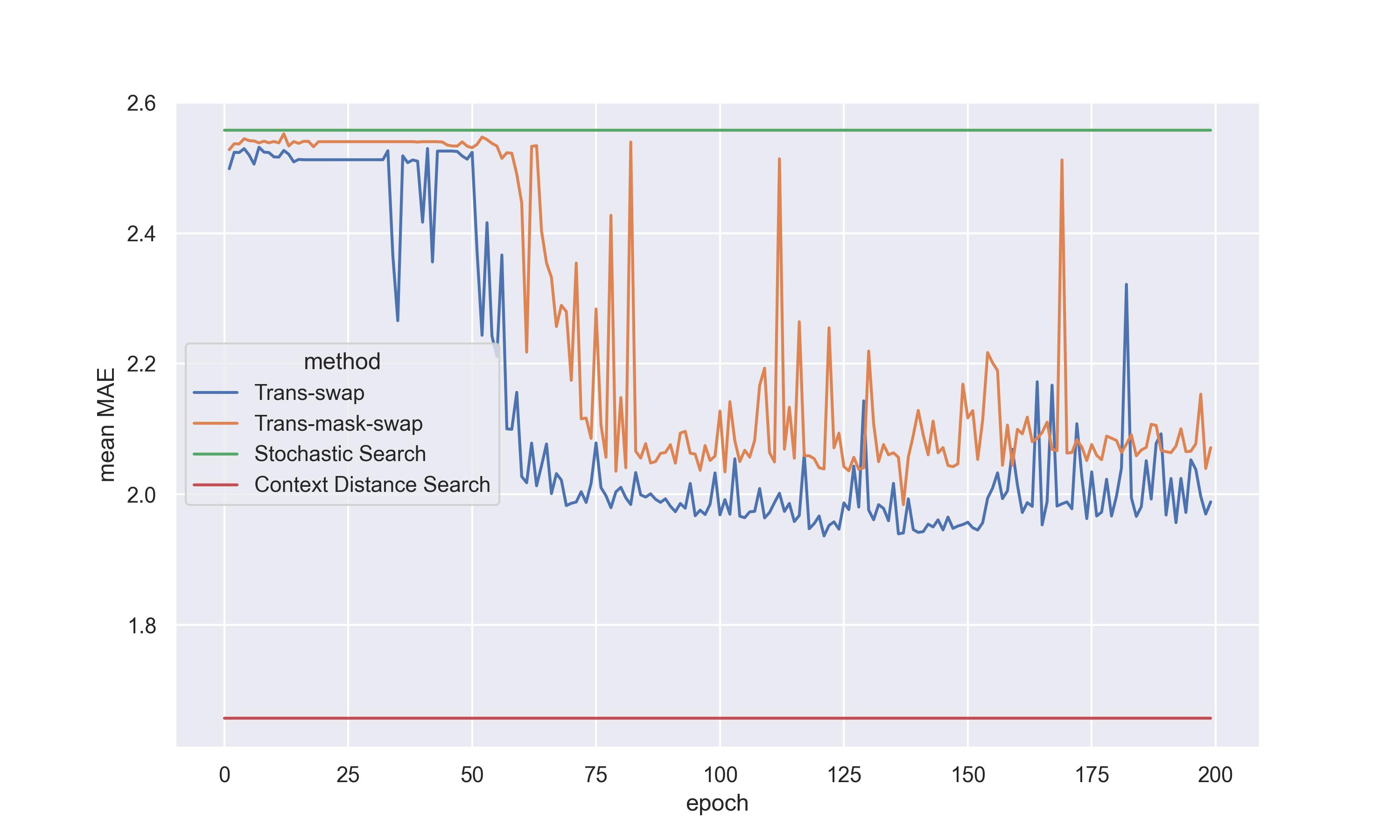}
		\caption{Mean of the mean MAE values on 60 \% testing instances} 
	\end{subfigure}
	\begin{subfigure}{0.95\columnwidth} 
		\includegraphics[width=\textwidth]{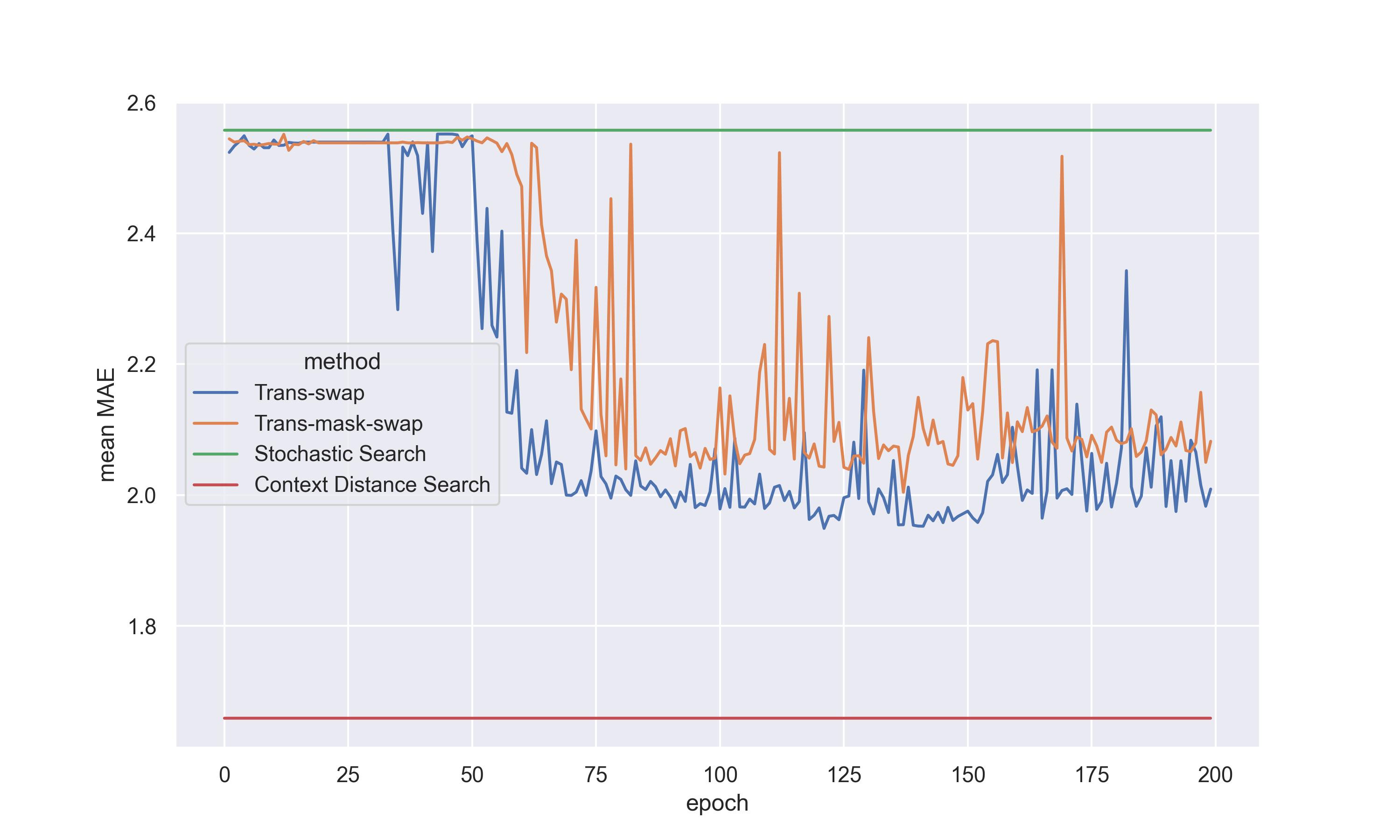}
		\caption{Mean of the mean MAE values on 80 \% testing instances} 
	\end{subfigure}
    \begin{subfigure}{0.95\columnwidth} 
		\includegraphics[width=\textwidth]{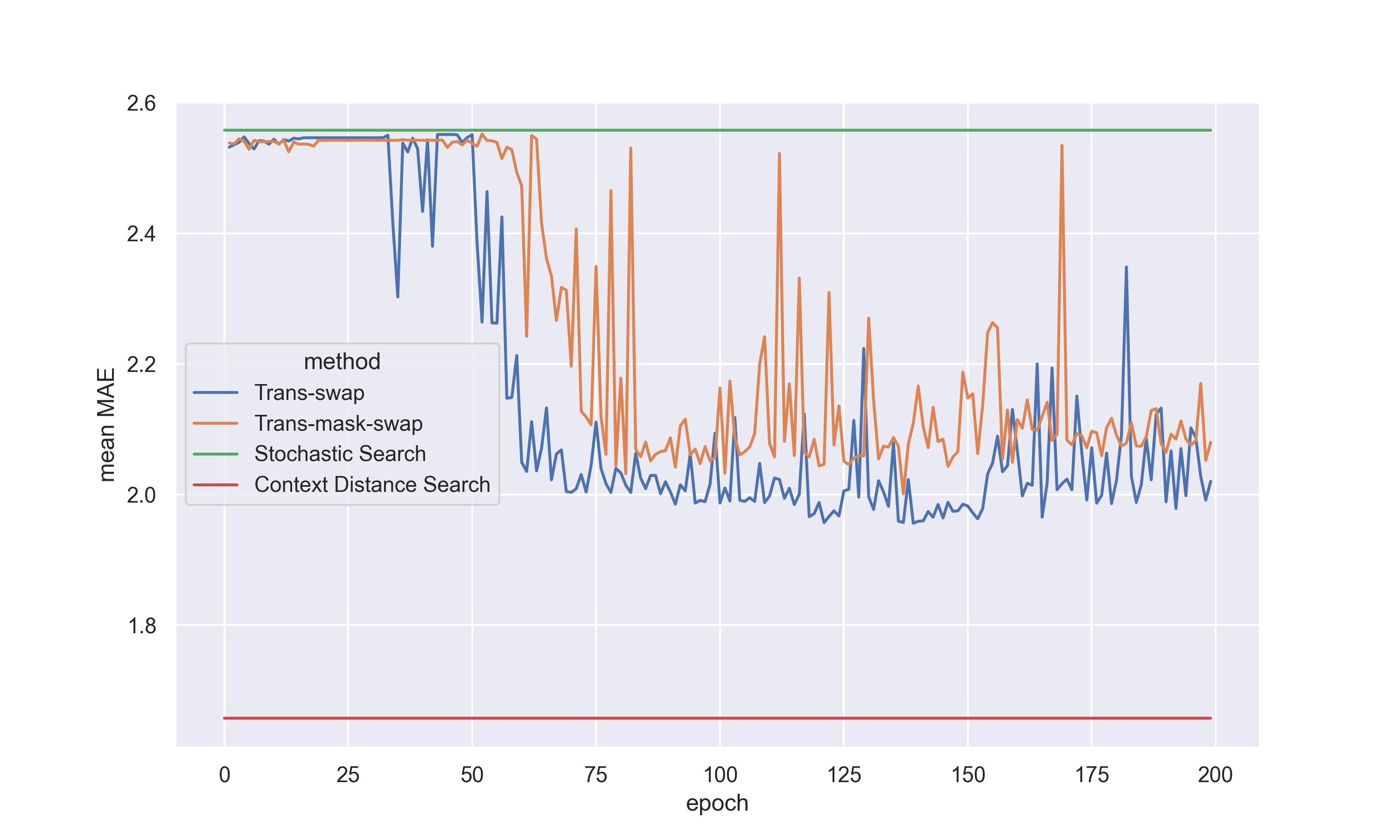}
		\caption{Mean of the mean MAE values on 100 \% testing instances} 
	\end{subfigure}
     \begin{subfigure}{0.95\columnwidth} 
		\includegraphics[width=\textwidth]{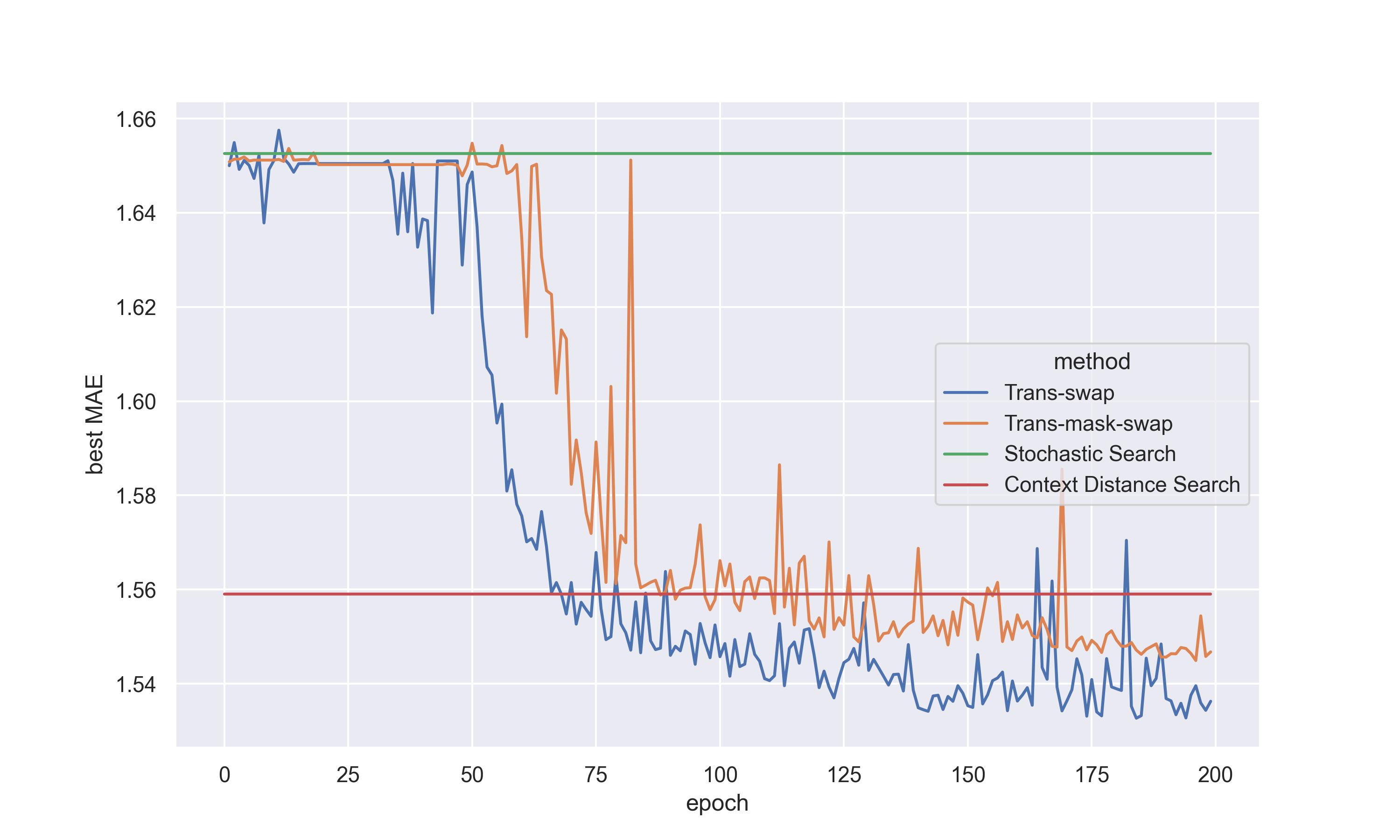}
		\caption{Mean of the best MAE values on 100 \% testing instances} 
	\end{subfigure}
\caption{Testing performance on trained policies over epochs}
\label{fig:testing_performance}
\end{figure*}

We first measure the mean of average MAE values over varying proportions of the 1000 instances, specifically, 20\%, 40\%, 60\%, 80\%, and 100\%. We conduct this test for each epoch from 0 to 199, enabling us to track the learning progress of the different methods over time. The results are plotted and presented in Fig.~\ref{fig:testing_performance}(a) - Fig.~\ref{fig:testing_performance}(e) for 20\%, 40\%, 60\%, 80\%, and 100\% testing instances, respectively. A clear pattern emerges from the plot: besides Context Distance Search, the `Tran-swap' method consistently outperforms the other techniques across all instances and epochs and for all percentages. This indicates that the `Tran-swap' method is more effective at reducing the average MAE, thereby producing more high-quality sensor placement solutions.

In addition to the mean MAE, we also evaluate the mean of the best MAE values obtained for the 1000 randomly generated instances, see Fig.~\ref{fig:testing_performance}(f). This measure gives us an indication of the best performance that each method can achieve. Our finding is not consistent with the previous results that Context Distance Search was the winner. the `Tran-swap' method emerges as the winner, and  it consistently achieves the lowest best MAE. In this paper, we are mainly interested in the best MAE for each problem instance, rather than mean MEA. Our findings demonstrates the superior performance of `Tran-swap' method in finding high-quality sensor placement solutions. This superior performance of our 'Tran-swap' method can be attributed to the combination of a transformer-based policy and reinforcement learning. The transformer’s attention mechanism effectively concentrates on critical sensor pairs for swapping, thereby enhancing search efficiency. Reinforcement learning, meanwhile, guides the policy towards maximizing the cumulative reward—finding optimal sensor placements. Over time, this approach refines the heuristic, leading to a lower best MAE, demonstrating the effectiveness of the transformer-based policy and reinforcement learning in improving heuristic solutions.

In summary, our proposed `Tran-swap' method exhibits excellent performance in both the average and best MAE measures, making it a promising approach for tackling the Sensor Placement Problem. While the `Tran-mask swap' method also shows good performance, it does not surpass the `Tran-swap' method in the tested scenarios.


\begin{table*}[ht]
\caption{The improvement on the mean and best MAE of four competing methods tested over 200, 400, 600, 800, and 1000 instances (Note: the lower the value the better)}
\label{tbl: scenarios}
\centering
\begin{tabular}{|c |c c c c|} 
\hline
\makecell{Problem instance \\ Number} & \makecell{Stochastic Search \\(mean value)} & \makecell{Context Distance Search \\ (mean value)} & \makecell{Trans-mask-swap \\(mean value)} & \makecell{Trans-swap \\(mean value)}  \\
\hline \hline
200 & 2.3991 $\pm$ 0.0149 &1.6558 $\pm$ 0.0634  &2.0397 $\pm$ 0.4095 &1.9060 $\pm$ 0.3840\\
400 & 2.5229 $\pm$ 0.0140 &1.6553 $\pm$ 0.0595 &2.0683 $\pm$ 0.4240 &1.9748 $\pm$ 0.4363\\
600 &2.5310  $\pm$ 0.0138 &1.6571 $\pm$ 0.0588 &2.0714 $\pm$ 0.4168 &1.9883 $\pm$ 0.4332 \\
800 &2.5445  $\pm$ 0.0140 &1.6589 $\pm$ 0.0584 &2.0822 $\pm$ 0.4069 &2.0091 $\pm$ 0.4419\\
1000 & 2.5577 $\pm$ 0.0139 &1.6574 $\pm$ 0.0596 &2.0801 $\pm$ 0.4093 &2.0199 $\pm$  0.4389\\ 
\hline
\hline
 \makecell{Problem instance \\ Number}  &\makecell{Stochastic Search \\(best value)} & \makecell{Context Distance Search \\(best value)} & \makecell{Trans-mask-swap \\(best value)} & \makecell{Trans-swap \\(best value)}  \\
\hline
1000 & 1.6525 $\pm$ 0.0053 & 1.5590 $\pm$ 0.0607 &1.5467 $\pm$ 0.1133 &1.5362 $\pm$ 0.1229\\
\hline
\end{tabular}
\end{table*}

\begin{figure}[!hbt]
\small
\centering
\begin{subfigure}{0.85\columnwidth} 
\includegraphics[width=\textwidth]{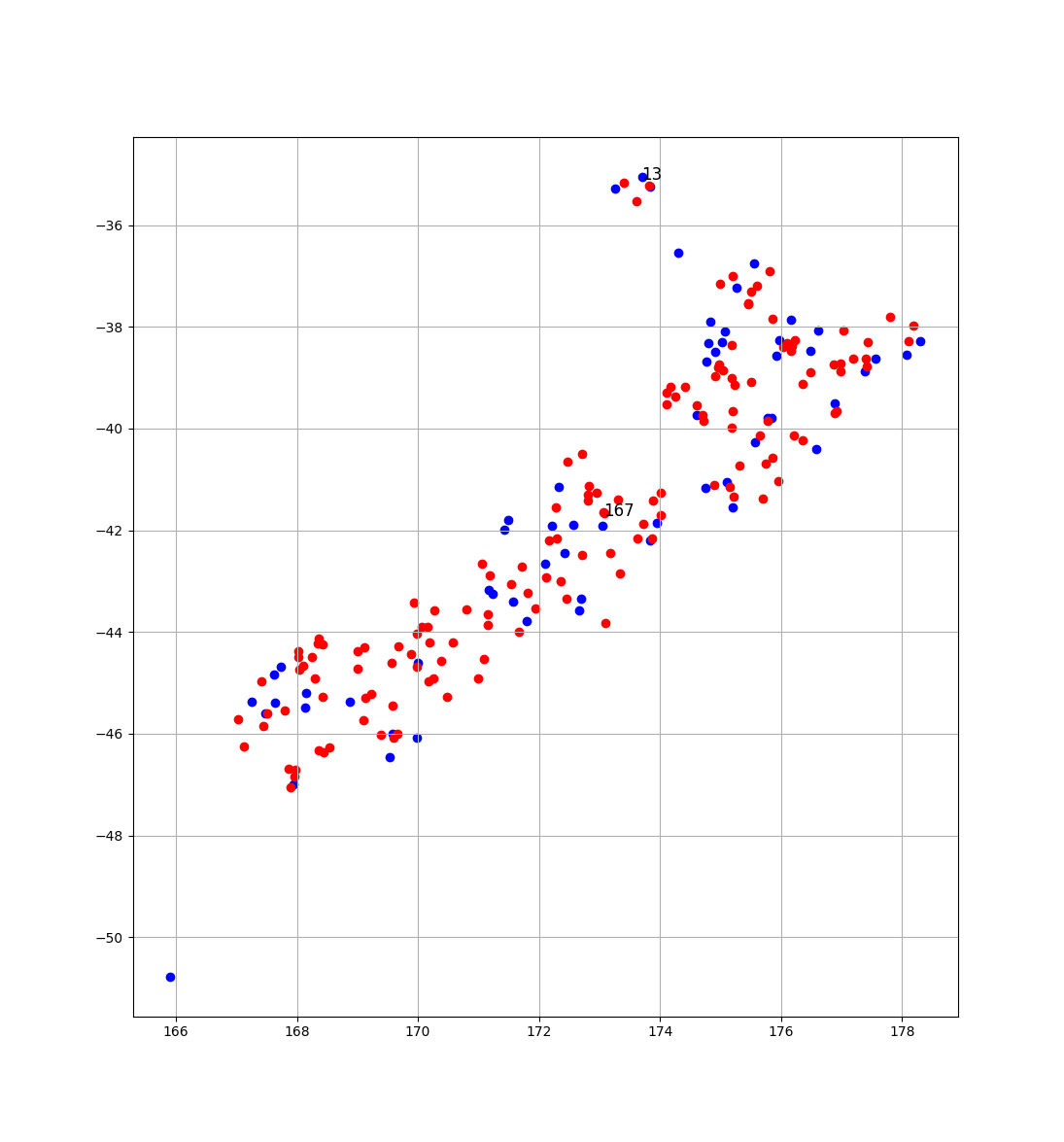}
\caption{An illustration of an initial placement solution: blue and red dots signify locations equipped with sensors and locations without sensors, respectively.} 
\end{subfigure}
\begin{subfigure}{0.95\columnwidth} 
\includegraphics[width=\textwidth]{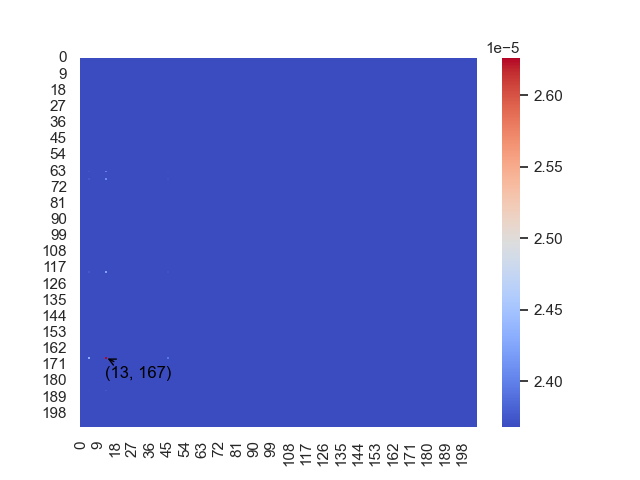}
\caption{An example of action table generated by the decoder} 
\end{subfigure}
\caption{Illustrating an initial solution as input, and explaining how our transformer policy can guide the operation of sensor movement.}
\label{fig:explain_model}
\end{figure}

\subsection{Parameters sensitivity}\label{subsection:parameters}
\subsubsection{Input embedding and hidden layer dimensions}
Choosing the correct dimensions for the input embedding and hidden layers is integral for the optimal performance of a Transformer-based model. We conducted an in-depth analysis to evaluate our model's sensitivity to these parameters. We tested three combinations of input embedding and hidden layer dimensions: (128, 128), (128, 256), and (256, 256). These specific combinations were chosen based on existing research \cite{wu2021learning}, which frequently use these dimensions due to their success in balancing model complexity and computational efficiency. Further, these dimensions have been found to provide a good trade-off between model complexity and performance.

\begin{figure}[!hbt]
\centering
	\begin{subfigure}{0.95\columnwidth} 
		\includegraphics[width=\textwidth]{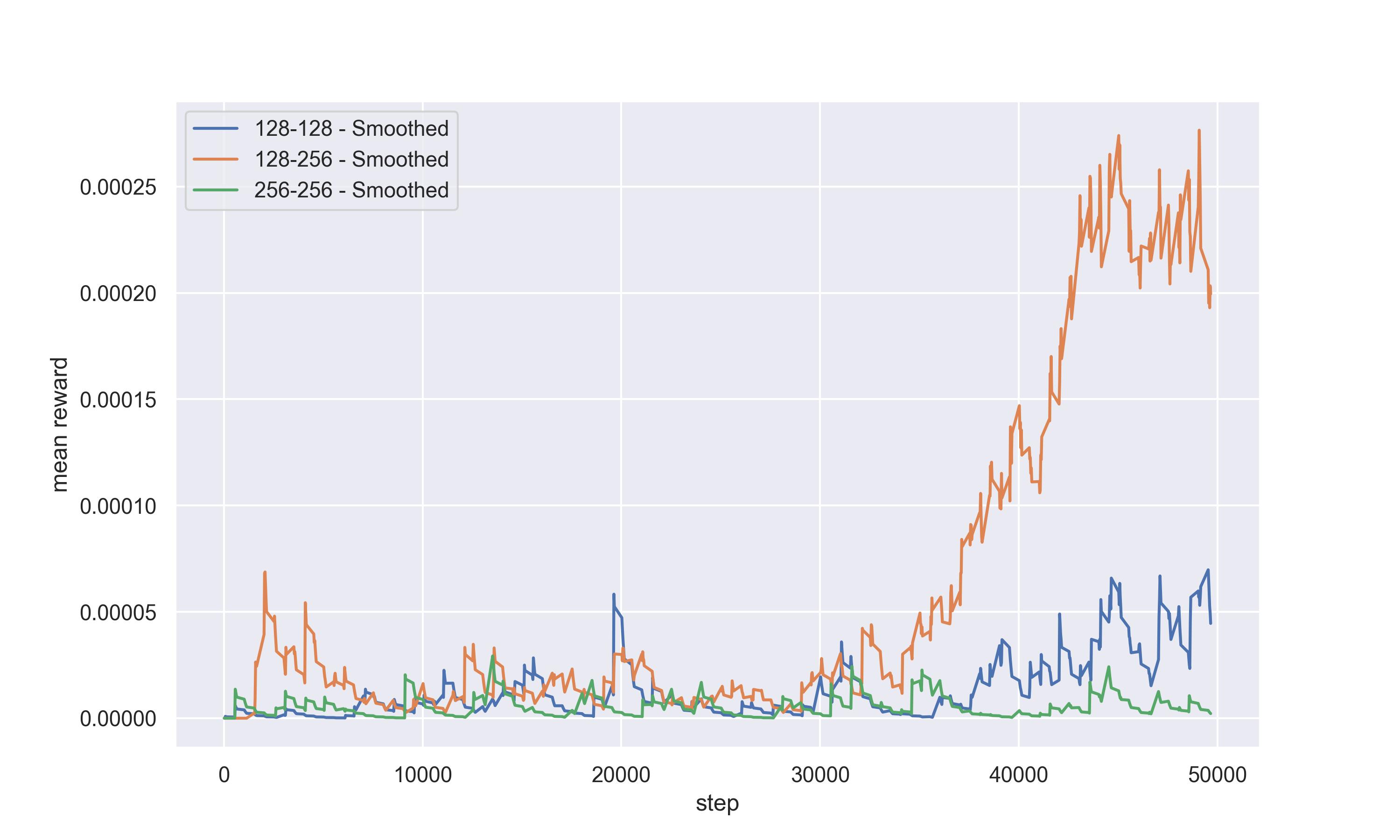}
		\caption{Mean of the mean rewards} 
	\end{subfigure}
	\begin{subfigure}{0.95\columnwidth} 
		\includegraphics[width=\textwidth]{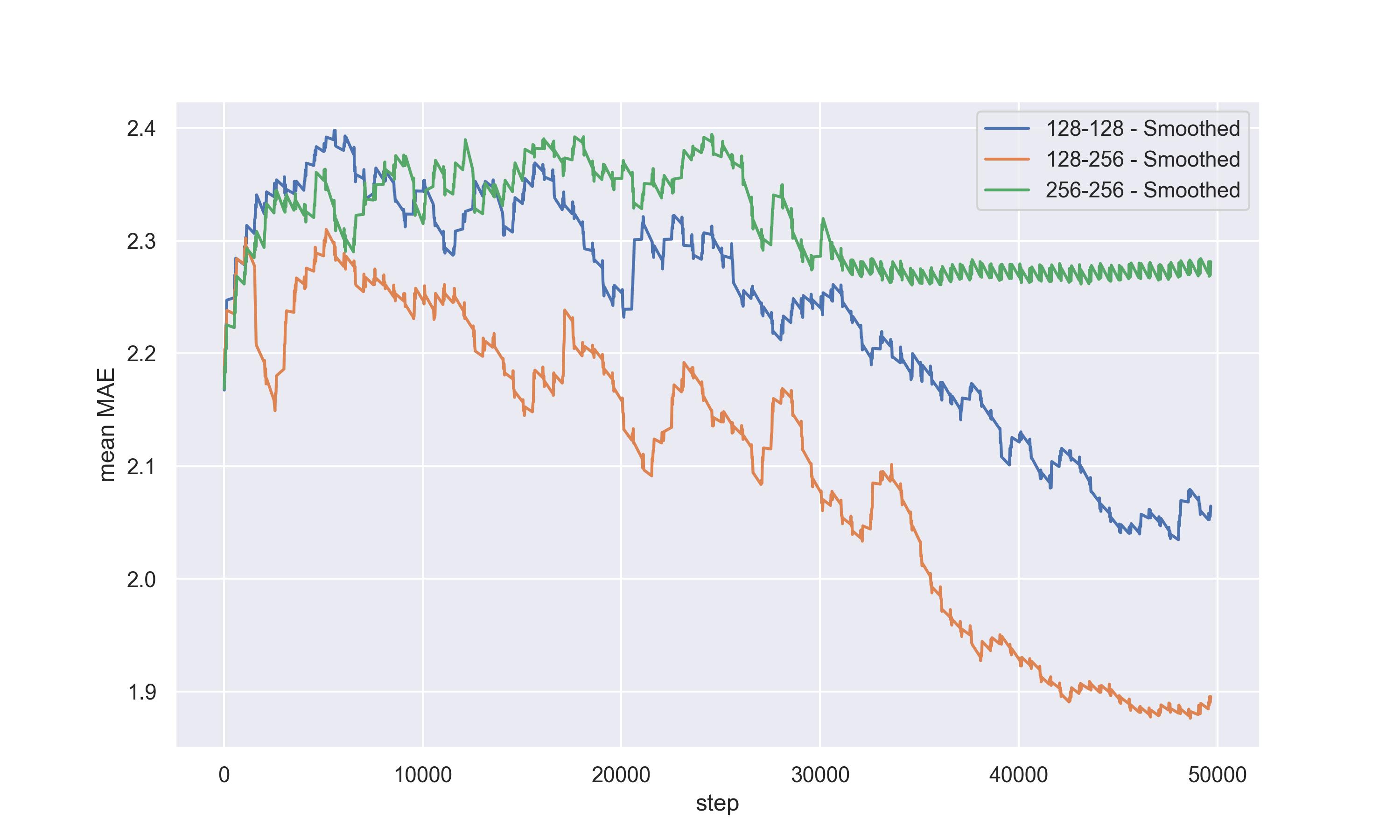}
		\caption{Mean of the mean MAE} 
	\end{subfigure}
\caption{Training performance of policies over steps using three different pairs of dimensions on input embedding and hidden layers}
\label{fig:sensitivity_dimension}
\end{figure}

\begin{figure}
\small
\centering
	\begin{subfigure}{0.95\columnwidth} 
		\includegraphics[width=\textwidth]{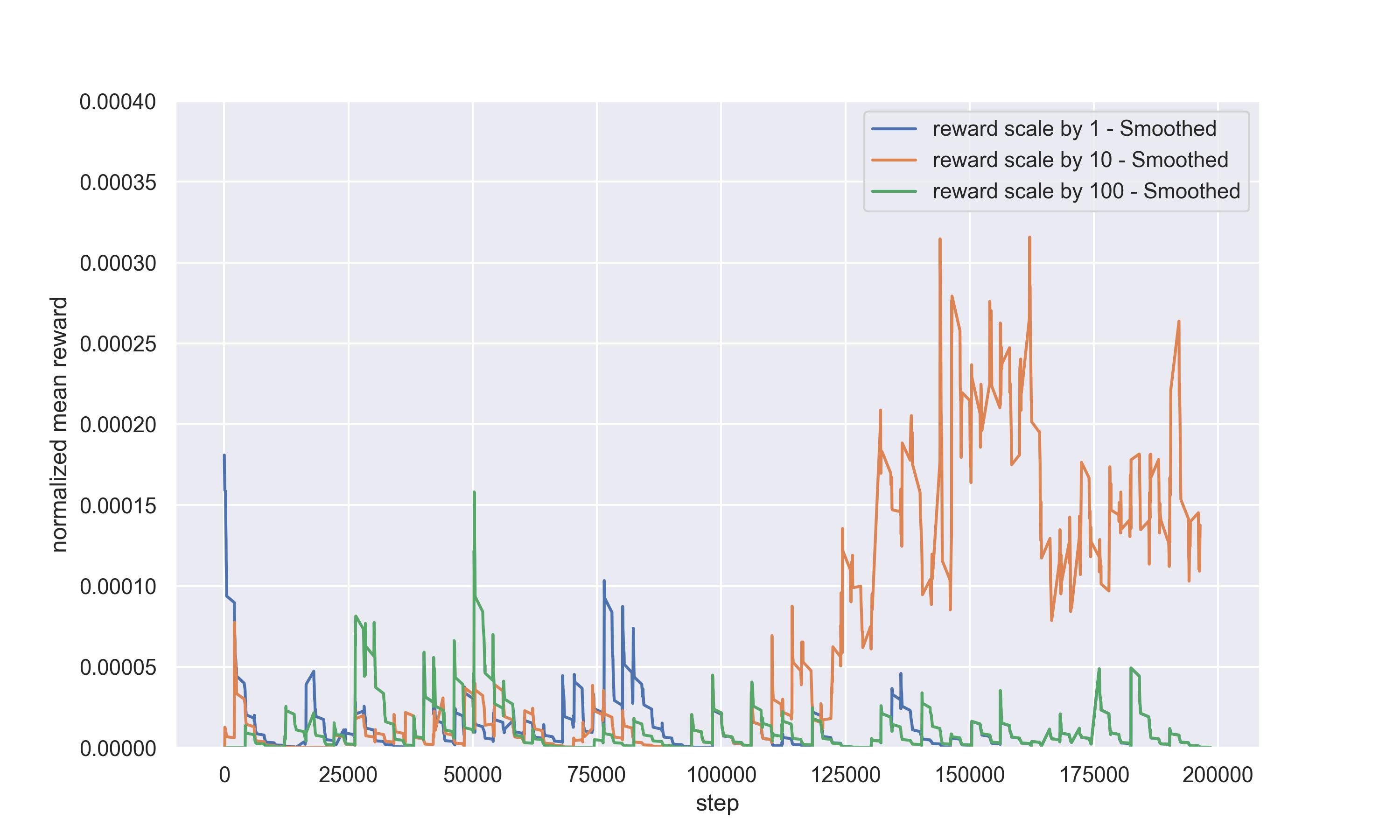}
		\caption{Mean of the mean rewards} 
	\end{subfigure}
	\begin{subfigure}{0.95\columnwidth} 
		\includegraphics[width=\textwidth]{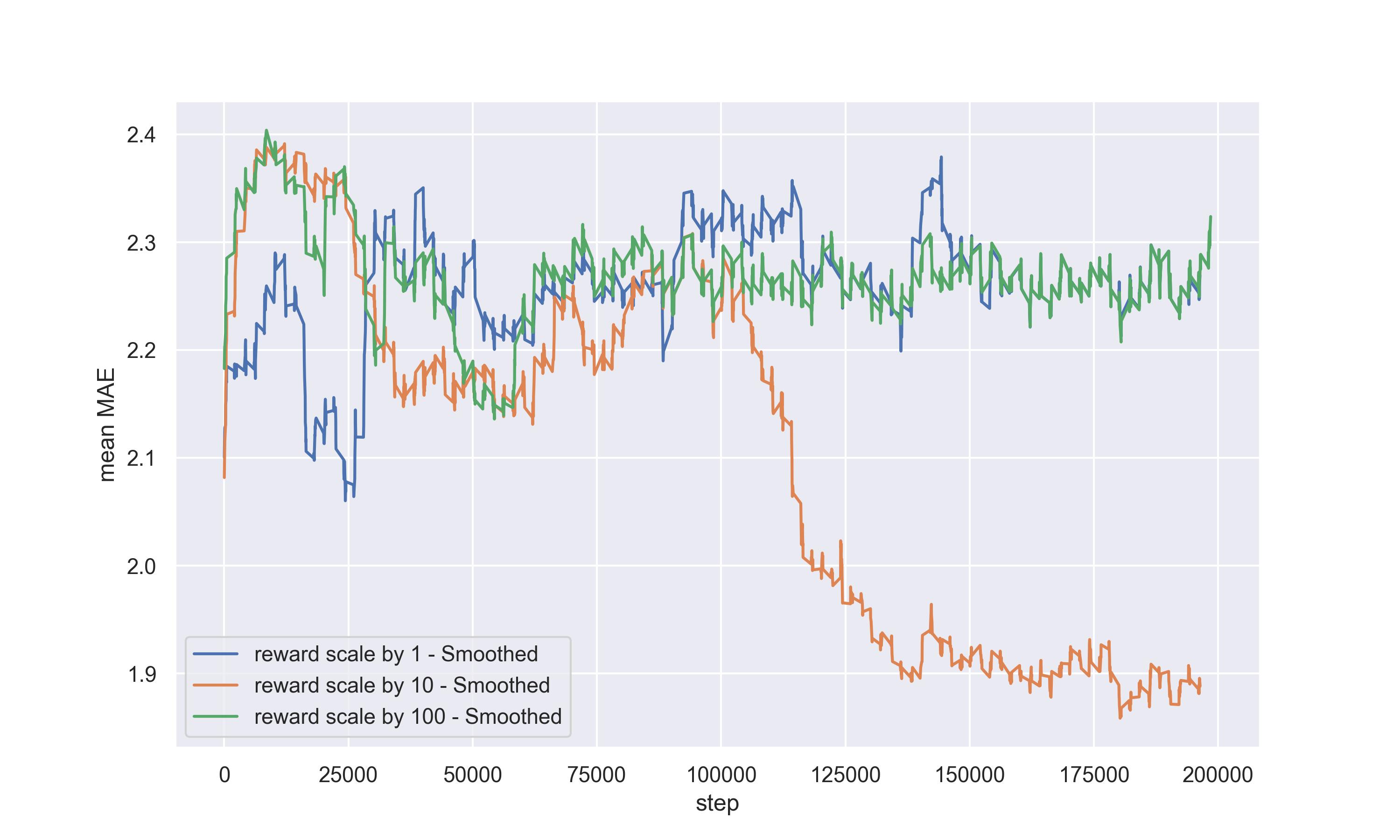}
		\caption{Mean of the mean MAE} 
	\end{subfigure}
\caption{Training performance of policies over steps using three different pairs layers}
\label{fig:train_perf}
\end{figure}

Fig.~\ref{fig:sensitivity_dimension}(a) and Fig.~\ref{fig:sensitivity_dimension}(b) visualize the learning curve of model performance over $50,000$ training steps using two key metrics: mean reward and mean MAE respectively. The (128, 256) configuration consistently achieved higher rewards and lower MAE, suggesting that a larger hidden layer dimension can improve performance, up to a point. Notably, increasing both dimensions to 256 did not enhance performance, hinting at a possible saturation effect or over-parameterization . These findings guide us towards optimal model configuration, balancing model complexity and performance.

\subsubsection{Reward scaling parameter}
The reward scaling parameter often influences the rate at which the policy weights are updated. An appropriate reward scaling parameter helps balance the trade-off between exploration and exploitation during the learning process \cite{yang2019never}. In the RL context, a larger reward scaling can encourage the agent to be greedy. However, by avoiding state and action pairs with lower rewards, the agent might not extensively explore the solution space, hindering them from gaining higher rewards in the long term. To determine the optimal reward scaling parameter, we conduct sensitivity analyses with three different scaling parameters of 1, 10, and 100. 

Fig.~\ref{fig:train_perf}(a) and Fig.~\ref{fig:train_perf}(b) show the performance trends for the mean reward and mean MAE across 200,000 steps, respectively. We can easily observe that the optimal reward scaling parameter is 10 which provides the highest mean of average reward and the lowest mean of MAE.

\section{Model Exploration} \label{sect:model_exploration}

In this section, we delve deeper into the complex details of our model to understand its internal functioning better. We begin this exploration by considering a randomly selected testing instance as an example. Fig.~\ref{fig:explain_model}(a) represents an initial solution to this testing instance, which is composed of a set of randomly determined sensor locations. The sensors are plotted on a 2D grid, which represents the geographical area under consideration for environmental monitoring. Each sensor location is denoted by a point in this grid.

\textbf{Passing the Input:}.
We pass this initial solution through the encoder-decoder architecture of our Transformer. The role of the encoder is to interpret the input, i.e., the initial sensor locations, and generate a high-dimensional representation that captures the essential features and relations of the input data. Then, the decoder uses this high-dimensional representation to generate an action table.

\textbf{Generating Action Table:}
The action table, visualized in Fig.~\ref{fig:explain_model}(b), represents the probable actions that our model suggests for the next step. Each action corresponds to a potential move of a sensor. The action table is a matrix where each entry represents the predicted reward of moving a specific sensor to a new location. Higher values in the table indicate a higher expected reward for the corresponding action. For example, we can see that by moving a sensor with location id 13 to a particular new location with id 167 achieves the highest reward as pointed out in Fig.~\ref{fig:explain_model}(b).

After generating the action table, we sample an action for the next step. We follow a stochastic policy for this selection: instead of always choosing the action with the highest expected reward, we sample an action from a probability distribution over the action space where the probability is proportional to the expected reward. With the help of a stochastic policy, we can better balance the exploration and exploitation.  

The chosen action then results in a new sensor configuration, which forms the input for the next iteration. This process is repeated until we reach a termination condition, such as a maximum number of iterations or a satisfactory solution quality. Through this mechanism, our model continually refines the sensor placement, guided by the policy it has learned via deep reinforcement learning. The result is a high-quality sensor configuration that has been adaptively optimized for the task at hand.

\section{Conclusions} \label{sect:conclusion}

In this paper, we have presented a novel sensor placement approach focused on learning improvement heuristics using deep reinforcement learning (RL) methods. This approach overcomes the limitations of traditional methods, such as exact methods, approximation methods, and heuristic methods, by automatically discovering effective improvement policies that can produce high-quality solutions. Our experimental results demonstrate the effectiveness and superiority of the proposed approach compared to state-of-the-art methods in solving the sensor placement problem.

Despite the promising results, there are several avenues for future research that could further improve the performance and applicability of our method. Some potential directions for future work include:

\begin{itemize}
\item \textbf{Teacher-student based reinforcement learning:} Incorporating a teacher-student learning paradigm, where a pretrained teacher network provides guidance to a student network during the training process, could help accelerate the learning process and improve the quality of the learned heuristics.

\item \textbf{Mixed learning:} Combining deep reinforcement learning with other learning techniques, such as supervised learning or unsupervised learning, may provide complementary benefits and enable our method to exploit a broader range of information during the learning process.


\end{itemize}

In conclusion, we believe that learning improvement heuristics using deep reinforcement learning offers a promising direction for solving complex optimization problems such as sensor placement. By continuing to explore and develop new techniques, models, and strategies, we can further enhance the capabilities of these methods in the future.
\bibliography{reference.bib}
\bibliographystyle{splncs04}
\begin{IEEEbiography}[{\includegraphics[width=1in,height=1.25in,clip,keepaspectratio]{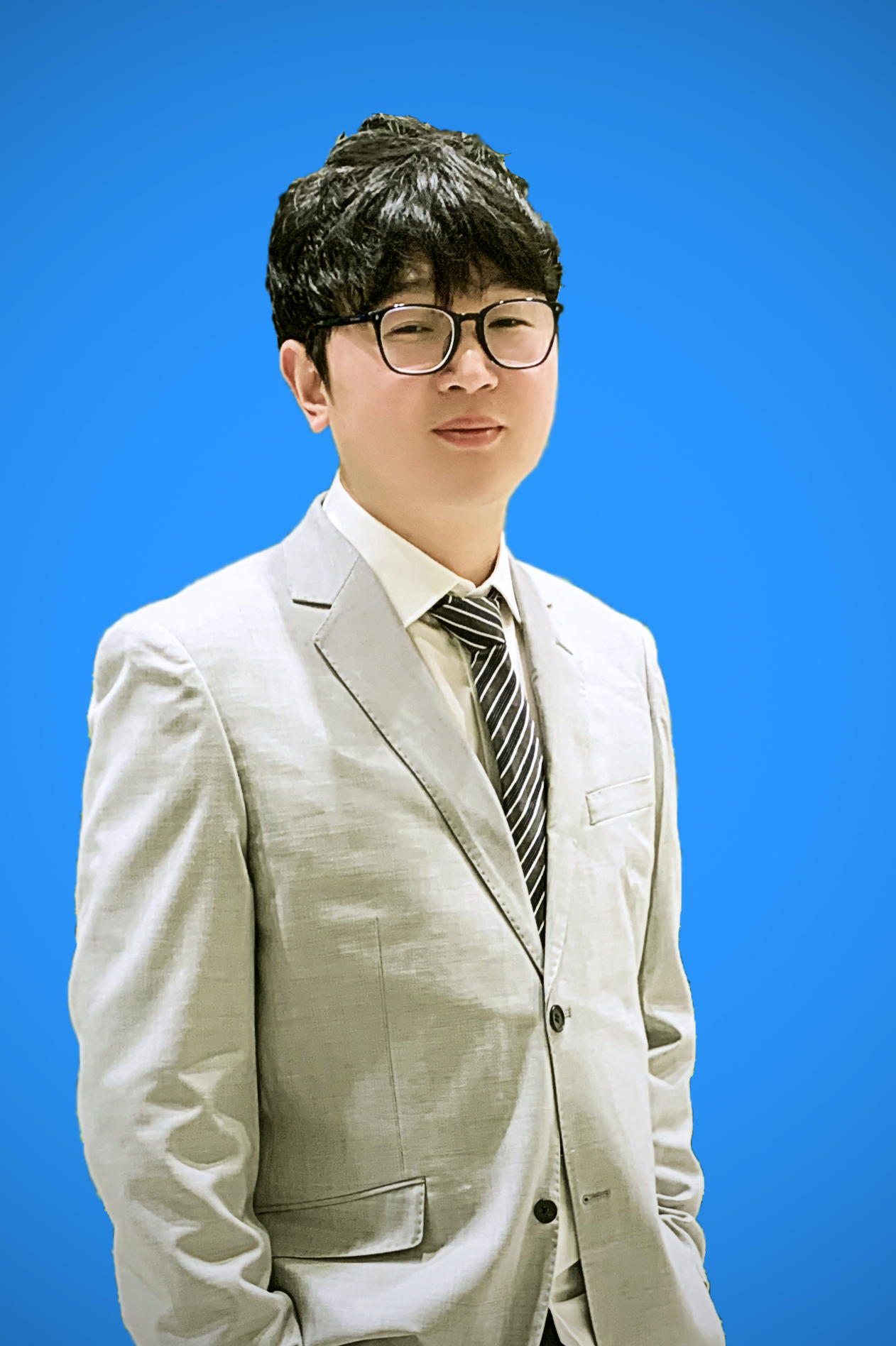}}]{Chen Wang} received his PhD degree in Engineering from Victoria University of Wellington, Wellington, New Zealand (2020). He is currently a data scientist at HPC and data science department from the National Institute of Water and Atmospheric Research, New Zealand. His research interests include optimisation and reinforcement learning techniques in solving challenging scientific problems on climate, fresh water and marine. \end{IEEEbiography}
\vspace{-1.5cm}
\begin{IEEEbiography}[{\includegraphics[width=1in,height=1.25in,clip,keepaspectratio]{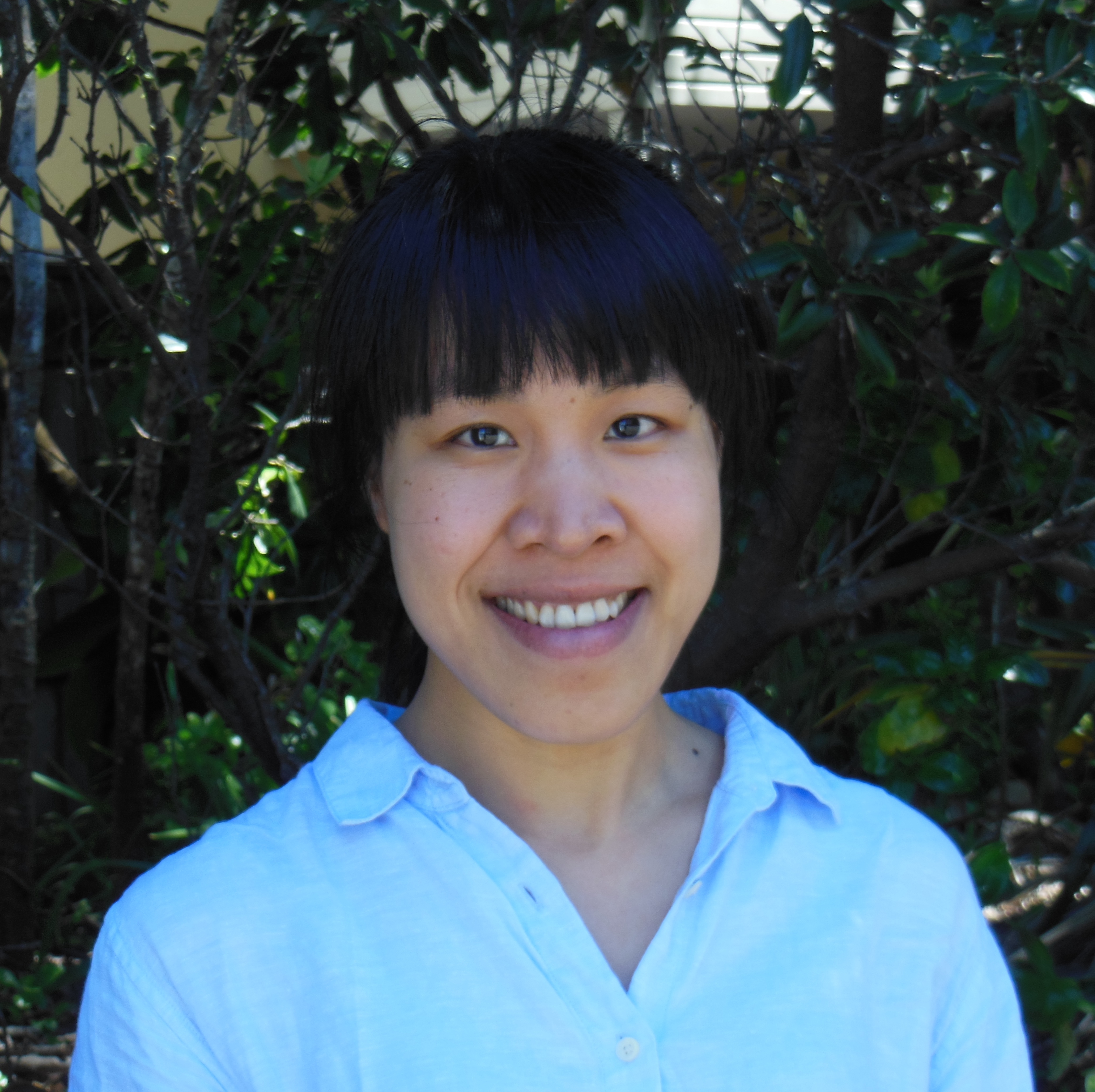}}]{Victoria Huang} received her Ph.D degree from Victoria University of Wellington, New Zealand. She is currently a data scientist in HPC and Data Science Department, National Institute of Water and Atmospheric Research, New Zealand. Her research interests include reinforcement learning, evolutionary computation algorithms, resource scheduling in Software-Defined Networking and cloud computing.\end{IEEEbiography}
\begin{IEEEbiography}[{\includegraphics[width=1in,height=1.25in,clip,keepaspectratio]{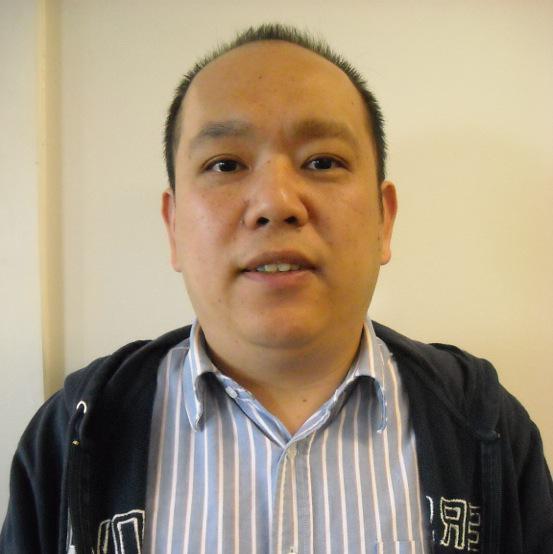}}]{Gang Chen} obtained his PhD degree from Nanyang Technological University (NTU) in 2007 in Singapore. He is currently a senior lecturer in the School of Engineering and Computer Science and Centre for Data Science and Artificial Intelligence (CDSAI) at Victoria University of Wellington. His research interests include evolutionary computation, reinforcement learning, multi-agent systems and cloud and service computing. He has more than 150 publications, including leading journals and conferences in machine learning, evolutionary computation, and distributed computing areas, such as IEEE TPDS, IEEE TEVC, JAAMAS, ACM TAAS, IEEE ICWS, IEEE SCC. He is serving as the PC member of many prestigious conferences including ICLR, ICML, NeurIPS, IJCAI, and AAAI, and co-chair for Australian AI 2018 and CEC 2019.\end{IEEEbiography}
\begin{IEEEbiography}[{\includegraphics[width=1in,height=1.25in,clip,keepaspectratio]{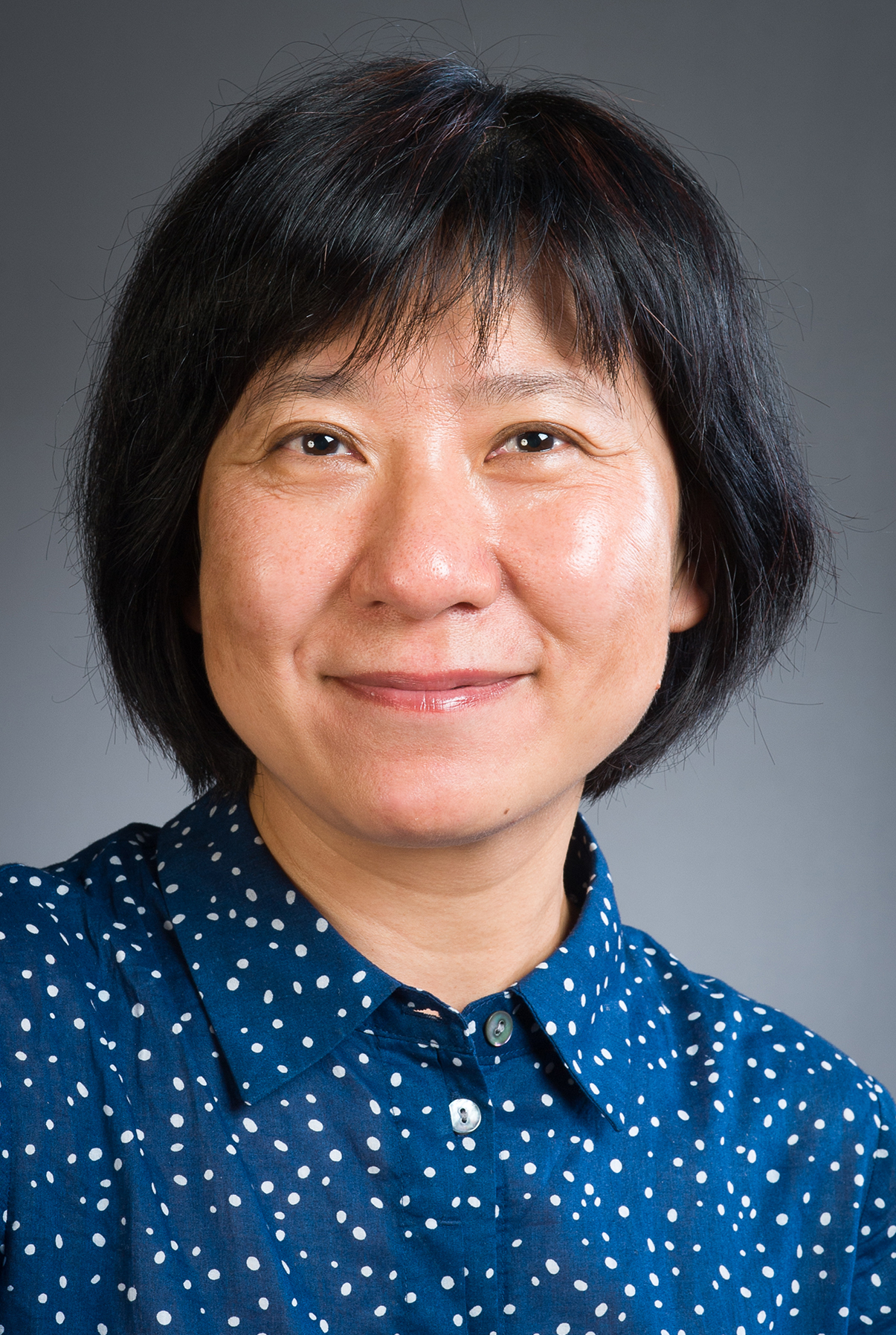}}]{Hui Ma} received her B.E. degree from Tongji University (1989) and her Ph.D degrees from Massey University (2008). She is currently an Associate Professor in Software Engineering at Victoria University of Wellington. Her research interests include service composition, resource allocation in cloud, conceptual modelling, database systems, resource allocation in clouds, and evolutionary computation in combinatorial optimization. Hui has more than 120 publications, including leading journals and conferences in databases, service computing, cloud computing, evolutionary computation, and conceptual modelling. She has served as a PC member for about 90 international conferences, including seven times as a PC chair for conferences such as ER, DEXA, and APCCM. \end{IEEEbiography}
\begin{IEEEbiography}[{\includegraphics[width=1in,height=1.25in,clip,keepaspectratio]{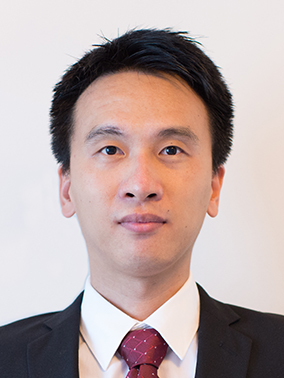}}]{Bryce Chen} graduated from Nanyang Technological Unversity, Singapore in 2008 with a Bachelor degree in Electrical and Electronics Engineering, major in Microelectronics. He obtained a Master of Science degree in Industrial Systems Engineering in 2015 from National University of Singapore. He is currently working as a data scientist in AL and DL team of HPC and Data Science Department, National Institute of Water and Atmospheric Research at New Zealand. His research interest includes time series related algorithms and application in anomaly detection and forecasting, as well as implementation of deep neural network algorithms in wide range of fields such as image recognition, object detection and natural language processing.\end{IEEEbiography}
\begin{IEEEbiography}[{\includegraphics[width=1in,height=1.25in,clip,keepaspectratio]{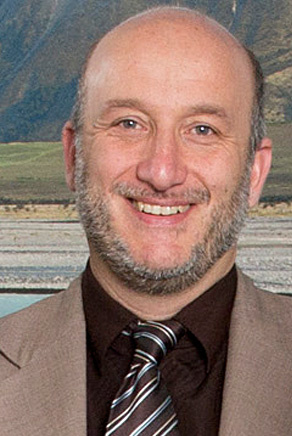}}]{Jochen Schmidt} received his PhD (Physical Geography), University of Bonn (2001) and Diploma (Geography) from University of Heidelberg (1996). Jochen has a background in hydrology, geomorphology, soil science, geo-informatics, and hazards and risk assessment. He worked for Landcare Research between 2001 and 2003 and was instrumental in developing the New Zealand Digital Soil Map (‘SMAP’). He joined NIWA as Chief Scientist - Environmental Information in 2003 and coordinates systems for collecting, managing and delivering environmental information – ensuring they are robust and meet best-practice standards.\end{IEEEbiography}
\end{document}